\documentclass[11pt]{article}

\usepackage[utf8]{inputenc}
\usepackage[T1]{fontenc}
\usepackage{mathptmx}
\usepackage[letterpaper,margin=1in]{geometry}
\usepackage{amsmath,amssymb}
\usepackage{graphicx}
\usepackage{booktabs}
\usepackage{tabularx}
\usepackage{multirow}
\usepackage{array}
\usepackage{threeparttable}
\usepackage{xcolor}
\usepackage{colortbl}
\usepackage[font=small,labelfont=bf,labelsep=colon]{caption}
\usepackage{enumitem}
\usepackage{microtype}
\usepackage{url}
\usepackage[hidelinks]{hyperref}
\usepackage{orcidlink}
\usepackage{twemojis}

\newcolumntype{L}{>{\raggedright\arraybackslash}X}
\newcolumntype{C}{>{\centering\arraybackslash}X}
\setlist{itemsep=2pt,topsep=4pt}

\definecolor{secbg}{HTML}{DEEAF6}
\definecolor{subbg}{HTML}{EFEFEF}

\DeclareRobustCommand\C[2]{\hyperlink{ref.#1}{#2}}

\expandafter\def\csname urpage@chor\endcsname{1}\expandafter\def\csname urdp@chor\endcsname{1.75pt}
\expandafter\def\csname urpage@nahin\endcsname{2}\expandafter\def\csname urdp@nahin\endcsname{3.40999pt}
\expandafter\def\csname urpage@hai\endcsname{3}\expandafter\def\csname urdp@hai\endcsname{4.51001pt}
\expandafter\def\csname urpage@kutte\endcsname{4}\expandafter\def\csname urdp@kutte\endcsname{2.81999pt}
\expandafter\def\csname urpage@kutta\endcsname{5}\expandafter\def\csname urdp@kutta\endcsname{0.0pt}
\expandafter\def\csname urpage@lanat\endcsname{6}\expandafter\def\csname urdp@lanat\endcsname{0.16998pt}
\expandafter\def\csname urpage@haram\endcsname{7}\expandafter\def\csname urdp@haram\endcsname{5.97998pt}
\expandafter\def\csname urpage@bara\endcsname{8}\expandafter\def\csname urdp@bara\endcsname{1.23999pt}
\expandafter\def\csname urpage@youthiyon\endcsname{9}\expandafter\def\csname urdp@youthiyon\endcsname{3.69998pt}
\expandafter\def\csname urpage@ho\endcsname{10}\expandafter\def\csname urdp@ho\endcsname{0.67998pt}
\expandafter\def\csname urpage@kisi\endcsname{11}\expandafter\def\csname urdp@kisi\endcsname{3.40999pt}
\expandafter\def\csname urpage@kabhi\endcsname{12}\expandafter\def\csname urdp@kabhi\endcsname{3.40999pt}
\expandafter\def\csname urpage@chutiya\endcsname{13}\expandafter\def\csname urdp@chutiya\endcsname{2.81998pt}
\expandafter\def\csname urpage@party\endcsname{14}\expandafter\def\csname urdp@party\endcsname{5.07999pt}
\expandafter\def\csname urpage@wohi\endcsname{15}\expandafter\def\csname urdp@wohi\endcsname{3.40999pt}
\expandafter\def\csname urpage@woh\endcsname{16}\expandafter\def\csname urdp@woh\endcsname{0.67998pt}
\expandafter\def\csname urpage@mein\endcsname{17}\expandafter\def\csname urdp@mein\endcsname{3.40999pt}
\expandafter\def\csname urpage@meri\endcsname{18}\expandafter\def\csname urdp@meri\endcsname{4.37999pt}
\expandafter\def\csname urpage@mera\endcsname{19}\expandafter\def\csname urdp@mera\endcsname{4.37999pt}
\expandafter\def\csname urpage@mujh\endcsname{20}\expandafter\def\csname urdp@mujh\endcsname{0.4pt}
\expandafter\def\csname urpage@lagta\endcsname{21}\expandafter\def\csname urdp@lagta\endcsname{0.0pt}
\expandafter\def\csname urpage@khud\endcsname{22}\expandafter\def\csname urdp@khud\endcsname{0.67998pt}
\expandafter\def\csname urpage@halanke\endcsname{23}\expandafter\def\csname urdp@halanke\endcsname{0.90999pt}
\expandafter\def\csname urpage@tha\endcsname{24}\expandafter\def\csname urdp@tha\endcsname{0.0pt}
\expandafter\def\csname urpage@tu\endcsname{25}\expandafter\def\csname urdp@tu\endcsname{0.67998pt}
\expandafter\def\csname urpage@ban\endcsname{26}\expandafter\def\csname urdp@ban\endcsname{3.40999pt}
\expandafter\def\csname urpage@balke\endcsname{27}\expandafter\def\csname urdp@balke\endcsname{1.87997pt}
\expandafter\def\csname urpage@baap\endcsname{28}\expandafter\def\csname urdp@baap\endcsname{4.99998pt}
\expandafter\def\csname urpage@apnichoti\endcsname{29}\expandafter\def\csname urdp@apnichoti\endcsname{3.40999pt}
\expandafter\def\csname urpage@sw1\endcsname{30}\expandafter\def\csname urdp@sw1\endcsname{0.90999pt}
\expandafter\def\csname urpage@sw2\endcsname{31}\expandafter\def\csname urdp@sw2\endcsname{3.40999pt}
\expandafter\def\csname urpage@sw3\endcsname{32}\expandafter\def\csname urdp@sw3\endcsname{0.0pt}
\expandafter\def\csname urpage@sw4\endcsname{33}\expandafter\def\csname urdp@sw4\endcsname{0.0pt}
\expandafter\def\csname urpage@sw5\endcsname{34}\expandafter\def\csname urdp@sw5\endcsname{3.40999pt}
\expandafter\def\csname urpage@sw6\endcsname{35}\expandafter\def\csname urdp@sw6\endcsname{0.67998pt}
\expandafter\def\csname urpage@sw7\endcsname{36}\expandafter\def\csname urdp@sw7\endcsname{3.40999pt}
\expandafter\def\csname urpage@sw8\endcsname{37}\expandafter\def\csname urdp@sw8\endcsname{4.37999pt}
\expandafter\def\csname urpage@sw9\endcsname{38}\expandafter\def\csname urdp@sw9\endcsname{0.4pt}
\expandafter\def\csname urpage@sw10\endcsname{39}\expandafter\def\csname urdp@sw10\endcsname{3.40999pt}
\expandafter\def\csname urpage@sw11\endcsname{40}\expandafter\def\csname urdp@sw11\endcsname{4.37999pt}
\expandafter\def\csname urpage@sw12\endcsname{41}\expandafter\def\csname urdp@sw12\endcsname{3.40999pt}
\expandafter\def\csname urpage@t4a1\endcsname{42}\expandafter\def\csname urdp@t4a1\endcsname{21.81999pt}
\expandafter\def\csname urpage@t4a2\endcsname{43}\expandafter\def\csname urdp@t4a2\endcsname{5.97998pt}
\expandafter\def\csname urpage@t4b1\endcsname{44}\expandafter\def\csname urdp@t4b1\endcsname{5.14998pt}
\expandafter\def\csname urpage@t4b2\endcsname{45}\expandafter\def\csname urdp@t4b2\endcsname{5.14998pt}
\expandafter\def\csname urpage@t4c1\endcsname{46}\expandafter\def\csname urdp@t4c1\endcsname{19.0pt}
\expandafter\def\csname urpage@t4c2\endcsname{47}\expandafter\def\csname urdp@t4c2\endcsname{25.75996pt}
\expandafter\def\csname urpage@t4d1\endcsname{48}\expandafter\def\csname urdp@t4d1\endcsname{28.95993pt}
\expandafter\def\csname urpage@t4d2\endcsname{49}\expandafter\def\csname urdp@t4d2\endcsname{28.95993pt}
\expandafter\def\csname urpage@t4e1\endcsname{50}\expandafter\def\csname urdp@t4e1\endcsname{5.07999pt}
\expandafter\def\csname urpage@t4e2\endcsname{51}\expandafter\def\csname urdp@t4e2\endcsname{5.07999pt}
\expandafter\def\csname urpage@t4f1\endcsname{52}\expandafter\def\csname urdp@t4f1\endcsname{27.22997pt}
\expandafter\def\csname urpage@t4f2\endcsname{53}\expandafter\def\csname urdp@t4f2\endcsname{27.22997pt}
\expandafter\def\csname urpage@s1\endcsname{54}\expandafter\def\csname urdp@s1\endcsname{6.612pt}
\expandafter\def\csname urpage@s2\endcsname{55}\expandafter\def\csname urdp@s2\endcsname{5.63998pt}
\expandafter\def\csname urpage@s3\endcsname{56}\expandafter\def\csname urdp@s3\endcsname{7.29599pt}
\expandafter\def\csname urpage@s4\endcsname{57}\expandafter\def\csname urdp@s4\endcsname{6.612pt}
\makeatletter
\newcommand\urpdf{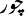}
\newcommand\ur[1]{\raisebox{-\csname urdp@#1\endcsname}{\includegraphics[page=\csname urpage@#1\endcsname]{\urpdf}}}
\newcommand\urb[1]{\par\hfill\includegraphics[page=\csname urpage@#1\endcsname]{\urpdf}\par}
\newcommand\urs[1]{\includegraphics[page=\csname urpage@#1\endcsname,width=\linewidth,height=1.2cm,keepaspectratio]{\urpdf}}
\makeatother

\title{\textbf{An Explainable System for Offensive Content Detection in Nastaliq Urdu: Corpus Curation, Transformer Benchmarking, and Word-Level Attribution}}
\author{Muhammad Shahid Iqbal Malik\,\orcidlink{0000-0001-8396-3344}\\[4pt]
\small School of Computing Sciences, Pak-Austria Fachhochschule, Institute of Applied Sciences and Technology,\\
\small Haripur 22620, Pakistan\\
\small\href{mailto:shahid.malik@paf-iast.edu.pk}{\texttt{shahid.malik@paf-iast.edu.pk}}}
\date{}

\begin{document}
\maketitle

\begin{abstract}
Automated moderation systems are deployed at scale, yet they remain opaque to the users and moderators affected by their decisions, and their coverage is almost entirely confined to English. For Urdu written in the Nastaliq script no such system exists, and the annotated resources it would be built on are drawn largely from Twitter/X, where character limits constrain how users write. Interpretability has received little attention, leaving it unclear what such models actually learn. This study addresses these gaps in three steps. First, we curate and publicly release a corpus of 7,604 Nastaliq Urdu Facebook comments, extending an earlier corpus by correcting text-level inconsistencies and re-examining the original labels. Second, we fine-tune four Urdu-capable transformers under a common grid-search protocol and compare them against twelve feature--classifier baselines on identical data partitions. The best configuration, Multilingual Representations for Indian Languages (MuRIL), reaches 87.05\% accuracy and 87.06\% macro F1, improving on the strongest baseline by 5.52 percentage points, a margin that exceeds the 95\% confidence interval on test accuracy; the two leading transformers differ by less than the test set can resolve, and performance is stable across five optimizers. Third, we compute word-level SHAP attributions adapted to right-to-left rendering. Attribution mass concentrates on high-frequency slurs but extends to syntactically neutral function words and to political vocabulary characteristic of the source pages, indicating that aggregate performance may rest partly on lexical and topical regularities rather than a generalizable representation of offensiveness.

\end{abstract}

\section{Introduction}\label{sec:1}
The rise of social media platforms has empowered individuals with unprecedented freedom of expression and opportunities for positive engagement. However, this openness also facilitates the unchecked spread of harmful and offensive content, which can have serious consequences for victims in online communication. In extreme cases, such content can incite unrest, provoke suicide attempts, or trigger acts of retaliation and violence (\C{hinduja2010}{Hinduja \& Patchin, 2010}). Offensive speech is generally defined as: ``Any form of opinion or expression that uses profane, degrading, or abusive language against an individual or community'' (\C{chen2017}{H. Chen, Mckeever, \& Delany, 2017}). Since there's so much information and people have different views, manual detection of OEs is inefficient, inciting the need for automated detection systems.

Despite efforts, current prevention and filtering mechanisms on social media platforms remain insufficient, often less effective to safeguard users from online harm (\C{jahan2021}{Jahan \& Oussalah, 2021}). Key challenges include the impracticality of manual tagging, language diversity, reliance on delayed user reports, and the complexity of capturing nuanced and context-sensitive language with NLP models. While significant research has been conducted in high-resource languages like English for hate speech (\C{mazari2024}{Mazari, Boudoukhani, \& Djeffal, 2024}) and OED (\C{ayetiran2024}{Ayetiran \& Özgöbek, 2024}), a large number of world languages remain underrepresented. A few exceptions include as Danish (\C{sigurbergsson2019}{Sigurbergsson \& Derczynski, 2019}), and Chinese (\C{xiao2024}{Xiao, Bouamor, \& Zaghouani, 2024}), etc. However, the prevalence of offensive speech in various communities highlights the importance of expanding detection systems to low-resource languages, such as Urdu.

Simultaneously, the AI research community has raised growing concerns over the black-box nature of DL and ML models, especially in sensitive domains such as content moderation (\C{rudin2022}{Rudin et al., 2022}). There is a pressing need for XAI to demystify model predictions and provide trustworthy, human-understandable explanations. Recently, a number of studies have focused on enhancing explainability in hate speech detection (\C{fetahi2025}{Fetahi et al., 2025}; \C{hashir2025}{Hashir \& Kim, 2025}; \C{hashmi2025}{Hashmi, Yildirim Yayilgan, Abomhara, \& Akerkar, 2025}). Similarly, some efforts have been made to develop explainable approaches for detecting offensive language (\C{nghiem2024}{Nghiem \& Daumé III, 2024}), abusive content (\C{ashraf2024}{Ashraf, Hosen, Asgar, Islam, \& Nawar, 2024}), and threatening speech (\C{malik2025}{M. S. I. Malik, Nazarova, \& Jamjoom, 2025}; \C{nazarova2024}{Nazarova, Malik, Ignatov, \& Hussain, 2024}). However, only a limited body of literature has addressed these explainability challenges in the context of under-resourced languages.

In recent years, research on automated systems for OED has begun to address the complexities of under-resourced languages, with efforts targeting Urdu (\C{hussain2022}{Hussain, Malik, \& Masood, 2022}), Roman Urdu (\C{akhter2020}{Akhter, Jiangbin, Naqvi, Abdelmajeed, \& Sadiq, 2020}), and Arabic (\C{mubarak2020}{Mubarak, Rashed, Darwish, Samih, \& Abdelali, 2020}). However, this shift introduces several new challenges, including linguistic complexity, contextual ambiguity, and a lack of annotated datasets. Focusing specifically on Urdu, four key limitations have been observed; First, most existing approaches rely on hand-crafted features (\C{raza2023}{Raza et al., 2023}), or semantic/linguistic cues (\C{akhter2020}{Akhter et al., 2020}; \C{humayoun2022}{Humayoun, 2022}), which often fail to capture the full context of offensive language. Second, to the best of our knowledge, interpretability has received limited attention in existing detection systems for offensive content in Nastaliq Urdu. Third, prior studies generally lack a comprehensive comparison between machine learning and deep learning models, limiting insight into optimal strategies. Finally, existing Nastaliq Urdu datasets often suffer from annotation inconsistencies that constrain model performance and generalization. These four limitations directly motivate the research objectives below:

\begin{enumerate}
\item To develop a high-performance, automated detection framework for offensive language in Nastaliq Urdu social media content.
\item To build an interpretable system that offers transparent, trustworthy, and understandable explanations of its predictions.
\end{enumerate}

\begin{table}[htbp]
\centering
\caption{Glossary of key terms}
\label{tab:1}
\small
\begin{tabularx}{\linewidth}{@{}l L l L@{}}
\toprule
\textbf{Term} & \textbf{Definition} & \textbf{Term} & \textbf{Definition} \\
\midrule
OE & Offensive Expressions & NLP & Natural Language Processing \\
DL & Deep Learning & ML & Machine Learning \\
XAI & Explainable Artificial Intelligence & AI & Artificial Intelligence \\
SHAP & Shapley Additive Explanations & LR & Logistic Regression \\
Bi-LSTM & Bidirectional Long Short-Term Memory & OED & Offensive Expressions Detection \\
SVM & Support Vector Machine & LSTM & Long Short-Term Memory \\
NB & Na\"ive Bayes & CNN & Convolutional Neural Network \\
BOW & Bags of words & ANN & Artificial Neural Network \\
DT & Decision Tree & BERT & Bidirectional Encoder Representations from Transformers \\
RoBERTa & Robustly Optimized Bidirectional Encoder Representations from Transformers & LIME & Local Interpretable Model-agnostic Explanations \\
AcME & Accelerated Model-agnostic Explanations & Twhin-BERT & Twitter Heterogeneous Information Network BERT \\
MuRIL & \href{https://arxiv.org/abs/2103.10730}{Multilingual Representations for Indian Languages} & & \\
\bottomrule
\end{tabularx}
\end{table}

This study aims to handle the above-mentioned challenges for OED in the Nastaliq Urdu language. We explored four state-of-the-art Urdu transformer models with fine-tuning hyper-parameters to design a robust identification system. The post-hoc SHAP XAI model is used to interpret the predictions of the classification process on the word level for individual instances. With this addition, the proposed framework provides meaningful and insightful interpretations of the prediction inference. In addition, a Nastaliq Urdu corpus is constructed by manually editing the comments to address inconsistency issues and revisiting the annotations of the Facebook corpus (\C{hussain2022}{Hussain et al., 2022}). Twelve baseline models are chosen to compare the performance of the proposed framework. The principal contributions of this study are as follows:

\begin{itemize}
\item \textbf{Curation of the system's knowledge base:} Extending the Facebook corpus of (\C{hussain2022}{Hussain et al., 2022}), we correct text-level inconsistencies and re-annotate every instance, changing 6.6\% of labels and raising inter-annotator agreement from 0.88 to 0.90. The resulting balanced set of 7,604 comments is released with full documentation.
\item \textbf{Word-level explanation of system decisions:} SHAP attributions concentrate on high-frequency slurs but extend to neutral function words and to political vocabulary from the source pages, indicating that aggregate performance may rest partly on lexical and topical shortcuts.
\item \textbf{A controlled transformer benchmark:} Four Urdu-capable transformers are fine-tuned under a common grid-search protocol and compared against twelve feature--classifier baselines on identical partitions, a control prior Urdu OED studies have not reported. Confidence intervals make explicit which margins the test partition can resolve.
\item \textbf{An assessment of optimization stability:} Performance is examined across five optimizers, with differences reported relative to the resolution of the test partition. MARS Shine-Through and Occlusion quantify the instances each classifier uniquely recovers or misses.
\end{itemize}

The remainder of the paper is described as; Related work is presented in section~\ref{sec:2}. Section~\ref{sec:3} presents the details of the proposed methodology in which all steps are described in detail. The description of the experimental setup is presented in section~\ref{sec:4}. Then, the experiments and their results are discussed in section~\ref{sec:5}. Section~\ref{sec:6} presents the discussion on results and limitations. The conclusion and future directions are presented in section~\ref{sec:7}. The list of abbreviations are presented in Table~\ref{tab:1}.

\section{Related Work}\label{sec:2}
In this section, a review of offensive text detection approaches in high and low-resource languages including Roman Urdu and Urdu is presented. The prior approaches are presented and compared in Table~\ref{tab:2}.

\subsection{OE Detection}\label{sec:2.1}
In 2017, a study (\C{davidson2017}{Davidson, Warmsley, Macy, \& Weber, 2017}) proposed a detection model that classifies tweets into hate speech, offensive, and neutral. The model used TF-IDF, Part of speech tags, and reading ease features and achieved 90\% F1-score with SVM. However, they did not address explainability of prediction inference. Later, Sigurbergsson and Derczynski (\C{sigurbergsson2019}{Sigurbergsson \& Derczynski, 2019}) proposed a detection model for offensive language and hate speech in the Danish language. The word embeddings, linguistic, and sentiment features are used with LR and Bi-LSTM. Their model achieved 70\% F1-score but left their framework as a black-box. Likewise, another study developed an OED model (\C{pitenis2020}{Pitenis, Zampieri, \& Ranasinghe, 2020}) for the Greek language. The TF-IDF (unigram and bigram) feature model is used with SVM, NB, and LSTM models. Their framework outperformed the baselines. Then a detection model is introduced by the study (\C{mubarak2020}{Mubarak et al., 2020}) for offensive content in the Arabic language. They utilized word embeddings with various ML models and achieved 83\% F1-score as the top performance but did not support explainability. Later, a Turkish dataset is introduced (\C{coltekin2020}{Çöltekin, 2020}) to identify offensive content. Word n-grams is combined with the SVM model and their framework outperformed the baselines. However, their framework is not explainable.

In 2020, an offensive language detection model in Roman Urdu and Urdu is proposed by the study (\C{akhter2020}{Akhter et al., 2020}). The word and char n-gram features are used with traditional ML models and got 95.9\% F1-score on Urdu corpora. However, their approach did not address explainability. Later, this work is extended by another model (\C{akhter2021}{Akhter, Jiangbin, Naqvi, AbdelMajeed, \& Zia, 2021}), focusing on the Urdu and Roman Urdu language for abusive content detection. They combined BOW with several ML and DL models. The conclusion is the outperformance of the CNN with 96.2\% accuracy on the Urdu corpus but without explainability. These figures are reported on different corpora, collected from different platforms and annotated under different schemes, and are therefore not directly comparable with results obtained on the corpus used in this study.

Then, an abusive content identification system (\C{amjad2021}{Amjad et al., 2021}) is proposed by using FastText embeddings, word and char n-grams features. In addition, they compared four traditional ML and two DL models and obtained the best results with the SVM model but did not address explainability of predictive inference. Another study addressed abusive and threatening content detection in the Urdu language (\C{kumar2021}{Kumar, Saumya, \& Roy, 2021}). They used an ensemble model and it obtained 81\% F1-score without explainability support. Then the study (\C{paaschcolberg2021}{Paasch-Colberg, Strippel, Trebbe, \& Emmer, 2021}) proposed a model to differentiate between various types of hate speech and offensive content in German. They conducted exhaustive experiments and demonstrated significant performance but did not address explainability. Another OEs detection model is proposed by the study (\C{hussain2022}{Hussain et al., 2022}). They explored word \& char n-grams, and word2vec features with an ensemble model and evaluated the performance on the Facebook corpus, which forms the basis of the corpus used in the present study. Their framework outperformed the baselines. However, their study left their framework as a black-box.

Then, Humayoun proposed a detection model for abusive and threatening text in Urdu (\C{humayoun2022}{Humayoun, 2022}). The BOWs and word2vec methods are used with SVM, AdaBoost, and CNN models. Their approach achieved 83.18\% F1-score for abusive text detection as a best performance but their approach did not facilitate explainability. Later, a hate speech and offensive language detection (\C{roy2022}{Roy, Bhawal, \& Subalalitha, 2022}) framework is developed for Malayalam and Tamil languages. The authors used word embeddings with ML and DL models and their detection model achieved 80\% and 93\% F1-scores for Malayalam and Tamil languages respectively. However their model missed explainability of predictive inference. A recent study (\C{saeed2023}{Saeed, Afzal, Rauf, \& Iltaf, 2023}) not only addressed the offensive and hate speech classification but also categorized the hate speech on severity basis. Their proposed framework used several ML and DL models with word2vec, and word n-gram features and obtained 86\% macro F1-score as the top-performance but did not address interpretability of classification inference. Another recent study proposed a combinatorial pre-processing approach for the detection of offensive content in the Urdu language (\C{raza2023}{Raza et al., 2023}). They explored TF-IDF with four ML models on Twitter and YouTube dataset and their model demonstrated better performance than baselines. However, their approach did not support explainability of predictive inference.

Although several of these studies evaluate multiple classifiers, none reports whether all models were trained and evaluated on identical data partitions, which limits the comparability of the figures they report.

\subsection{Explainable AI Approaches}\label{sec:2.2}
XAI approaches are being used to obtain meaningful and understandable interpretations of predictions generated by ML or DL models (\C{ali2023b}{S. Ali et al., 2023}). Literature suggests that a prediction model is defined as trustworthy if it satisfies interpretability, stability, fairness, and robustness criteria. The XAI techniques are mainly classified into four types; 1) Inherently explainable, 2) Hybrid explainable, 3) Post-hoc explainable, and 4) Joint predictable and explainable.

The inherently explainable models are initially designed with the facility of interpretability, and they are often called white-boxes. Logistic regression, decision tree, and k-nearest neighbor are few examples (\C{daud2015}{Daud, Ahmad, Malik, \& Che, 2015}). Some extended versions are rule-based techniques (\C{jung2017}{Jung, Concannon, Shroff, Goel, \& Goldstein, 2017}), clustering-based methods (\C{saisubramanian2020}{Saisubramanian, Galhotra, \& Zilberstein, 2020}), and hybridization of ANN and DT (\C{schmitz1999}{Schmitz, Aldrich, \& Gouws, 1999}). The second type of XAI models (hybrid category) employ an explainable approach along with a black-box model. Therefore, these models support better evaluation metrics compared to inherently XAI techniques. The examples are DNN hybridization with KNN (\C{papernot2018}{Papernot \& McDaniel, 2018}), and the combination of DNN with a symbolic knowledge base (\C{bennetot2022}{Bennetot, Franchi, Del Ser, Chatila, \& Diaz-Rodriguez, 2022}). The third type of explainable technique is post-hoc: these approaches use additional algorithms to interpret the predictions generated by the primary model. The primary model generates some useful information during the training process that is used by the additional model for the interpretation of results. Some well-known instances are AcME (\C{dandolo2023}{Dandolo, Masiero, Carletti, Dalle Pezze, \& Susto, 2023}), LIME (\C{ribeiro2016}{Ribeiro, Singh, \& Guestrin, 2016}), and SHAP (\C{lundberg2017}{Lundberg \& Lee, 2017}) algorithms. However, they have drawbacks of being computationally slow and exposed to manipulation in the results. The last category of XAI techniques adopts the process of explicit training to explain the predictions. These are kinds of customized development, some techniques are loss-based approach (\C{zhang2018}{Q. Zhang, Wu, \& Zhu, 2018}), prototype identification (\C{chen2019}{C. Chen et al., 2019}), etc.

\begin{table}[htbp]
\centering
\caption{Summary of offensive/abusive content detection approaches in high and low resource languages}
\label{tab:2}
\footnotesize
\begin{threeparttable}
\begin{tabularx}{\linewidth}{@{}>{\raggedright\arraybackslash}p{2.6cm} >{\raggedright\arraybackslash}p{2.3cm} L >{\raggedright\arraybackslash}p{2.6cm} >{\raggedright\arraybackslash}p{2.6cm}@{}}
\toprule
\textbf{Year [Ref]} & \textbf{Platform} & \textbf{Feature Engineering} & \textbf{Classification} & \textbf{Algorithms} \\
\midrule
2017 (\C{davidson2017}{Davidson et al., 2017}) & English [Twitter] & TF-IDF, Part of speech, Reading ease scores & Hate speech and offensive content & SVM, LR \\
2019 (\C{sigurbergsson2019}{Sigurbergsson \& Derczynski, 2019}) & English \& Danish [Reddit and Facebook] & Linguistic, sentiment, reading ease, pre-trained embeddings & Offensive vs not-offensive & LR, Bi-LSTM \\
2020 (\C{coltekin2020}{\c{C}\"oltekin, 2020}) & Turkish [Twitter] & Word n-grams & Offensive vs not-offensive & SVM \\
2020 (\C{pitenis2020}{Pitenis et al., 2020}) & Greek [Twitter] & TF-IDF unigram and bigrams & Offensive vs not-offensive & SVM, NB, LSTM \\
2020 (\C{mubarak2020}{Mubarak et al., 2020}) & Arabic [Twitter] & Word embeddings & Offensive vs not-offensive & DT, RF, SVM, LR \\
2020 (\C{akhter2020}{Akhter et al., 2020}) & Urdu \& Roman Urdu [YouTube] & Word n-grams, char n-grams & Offensive vs not-offensive & NB, KNN, RF, SVM \\
2021 (\C{akhter2021}{Akhter et al., 2021}) & Urdu \& Roman Urdu [YouTube] & Bag of words & Abusive vs not-abusive & NB, SVM, CNN, LSTM \\
2021 (\C{amjad2021}{Amjad et al., 2021}) & Urdu [Twitter] & Word \& char n-grams, FastText & Abusive vs not-abusive & SVM, LR, RF, CNN, LSTM, AdaBoost \\
2021 (\C{kumar2021}{Kumar et al., 2021}) & Urdu [Twitter] & Char n-grams & Abusive and threatening task & SVM, Ensemble \\
2021 (\C{paaschcolberg2021}{Paasch-Colberg et al., 2021}) & German [News Articles] & Frequency-based features & Hate speech and offensive content & -- \\
2022 (\C{hussain2022}{Hussain et al., 2022}) & Urdu [Facebook] & Word \& char n-grams, word2vec & Offensive vs not-offensive & SVM, RF, LR, SGD, Ensemble model \\
2022 (\C{humayoun2022}{Humayoun, 2022}) & Urdu [Twitter] & Bag of words, word2vec & Abusive and threatening task & SVM, AdaBoost, CNN \\
2022 (\C{roy2022}{Roy et al., 2022}) & Tamil \& Malayalam [Twitter] & Word embeddings & Hate Speech and offensive language & CNN, LSTM \\
2023 (\C{saeed2023}{Saeed et al., 2023}) & Urdu [Twitter] & Word \& char n-grams, word and BERT embeddings & Offensive vs not, its severity & SVM, LR, CNN, LSTM, BERT \\
2023 (\C{raza2023}{Raza et al., 2023}) & Urdu [Twitter, YouTube] & TF-IDF & Offensive vs not-offensive & DT, RF, NB, LR \\
\midrule
\textbf{Proposed*} & Urdu [Facebook] & Word2vec, BERT, Twhin-BERT, RoBERTa, MuRIL with fine-tuning & Offensive vs non-offensive (with explainability) & BERT, Twhin-BERT, RoBERTa, MuRIL, CNN, SVM, RF, LR \\
\bottomrule
\end{tabularx}
\begin{tablenotes}[flushleft]\scriptsize
\item NB: Na\"ive Bayes; KNN: K-Nearest Neighbor; SVM: Support Vector Machine; RF: Random Forest; CNN: Convolutional Neural Network; LSTM: Long Short-Term Memory; DT: Decision Tree; LR: Logistic Regression; SGD: Stochastic Gradient Descent; Bi-LSTM: Bidirectional Long Short-Term Memory.
\end{tablenotes}
\end{threeparttable}
\end{table}

The literature review reveals following persistent limitations in OED research for Urdu:

\begin{itemize}
\item \textbf{Lack of research on Nastaliq Urdu}: limited research work addresses offensive content identification in this script.
\item \textbf{Limited interpretability}: Interpretability has received limited attention in existing OED work for Nastaliq Urdu.
\item \textbf{Limited feature engineering:} Prior approaches rely on lexical, frequency-based, and syntactic features rather than learned contextual representations.
\item \textbf{Lack of systematic model comparison:} Very few studies benchmark ML against DL models to identify the best-performing approach.
\end{itemize}

This study addresses each of these limitations directly. To overcome the reliance on hand-crafted features, we fine-tune four state-of-the-art transformer models that learn contextual representations of offensive Urdu language automatically. To address the interpretability gap, we integrate the post-hoc SHAP method to generate word-level explanations of model predictions. To enable systematic comparison, we benchmark our fine-tuned transformers against twelve standard ML and DL baselines. Finally, to address annotation quality issues, we manually revised the Facebook Nastaliq Urdu corpus (\C{hussain2022}{Hussain et al., 2022}) to correct inconsistencies and improve label reliability.

\section{Proposed Methodology}\label{sec:3}
\begin{figure}[!htbp]
\centering
\includegraphics[width=1.0\linewidth]{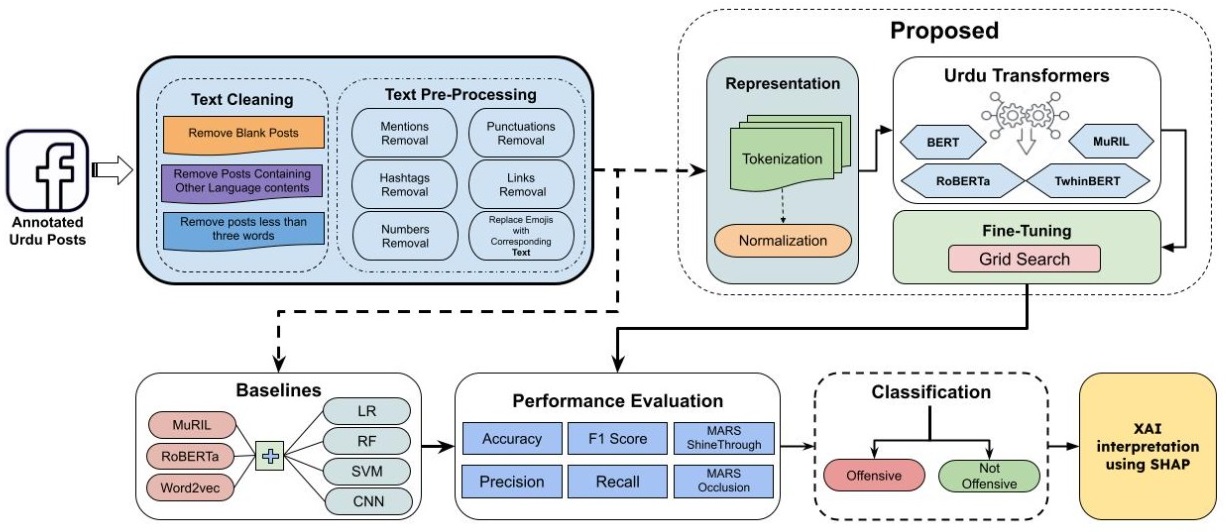}
\caption{Working flow of Explainable OED in Nastaliq Urdu}
\label{fig:1}
\end{figure}

In this study, we present an interpretable binary classification framework for OED in Nastaliq Urdu Facebook posts. The primary objective is to design a system that relies on automated feature extraction rather than hand-crafted features, while effectively capturing the contextual nuances of offensive language in Nastaliq Urdu with explainability. The overall architecture of the whole process is illustrated in Fig.~\ref{fig:1}. For the experimental setup, we employed a curated refined version of the Facebook Urdu corpus. The dataset was preprocessed using multiple distinct techniques to handle noise, tokenization, normalization, and script-related challenges. We evaluated four state-of-the-art transformer models using fine-tuning hyperparameters, among which MuRIL demonstrated the best performance. To benchmark our results, we compared MuRIL against twelve baseline models, including both classical machine learning and transformer-based approaches. To enhance interpretability, we integrated the SHAP framework, enabling word-level explanations of MuRIL's predictions.

Additionally, we applied two MARS-based evaluation metrics to assess the distinct strengths of the proposed framework. In summary, the proposed system is composed of robust preprocessing, MuRIL fine-tuning, and SHAP-based interpretability.

\subsection{Problem Formulation}\label{sec:3.1}
OED is a binary classification task, the formal formulation presented below:

Given a collection of Facebook posts $(F_1,$ $F_2, \ldots, F_n)$ and corresponding vectors $\{(X_1, Z_1),$ $(X_2, Z_2),$ $\ldots,$ $(X_n, Z_n)\}$, where $n$ represents the total number of posts; $X_i$ represents the corresponding feature vector of the post $F_i$ ($X_i \in \mathbb{R}^T$, $T$ refers to the total number of features), and $Z_i \in \{\text{offensive},$ $\text{non-offensive}\}$. To classify whether a Facebook post $F_i$ is offensive or non-offensive, a predictive function is defined:
\begin{equation}
Z_i = F_{\mathrm{OED}}\!\left(\frac{F_i}{X_i}\right)
\label{eq:1}
\end{equation}
where
\begin{equation}
F_{\mathrm{OED}}\!\left(\frac{F_i}{X_i}\right) =
\begin{cases}
> 0, & \text{if } Z_i = 1, \quad \text{Offensive} \\
= 0, & \text{if } Z_i = 0, \quad \text{Non-Offensive}
\end{cases}
\label{eq:2}
\end{equation}

\textbf{Objective function:} The goal here is to learn a predictive function that helps us to categorize whether a post is offensive or non-offensive.

\subsection{Mathematical Modelling of Semantic Model in Urdu}\label{sec:3.2}
Here we provide the mathematical description of the semantic model for the Urdu language. Usually, the semantic models use the Skip-gram architecture. Given a sequence of Urdu words $w_1, w_2, \ldots, w_T$, the objective is to maximize the average log probability:
\begin{equation}
\mathcal{L} = \frac{1}{T} \sum_{t=1}^{T} \sum_{\substack{-c \leq j \leq c \\ j \neq 0}} \log p\left(w_{t+j} \mid w_t\right)
\label{eq:3}
\end{equation}
where $c$ is the context window size. The probability $p(w_O \mid w_I)$ is defined using the Softmax function:
\begin{equation}
p\left(w_O \mid w_I\right) = \frac{\exp\left({V_{w_O}}^{\top} V_{w_I}\right)}{\sum_{w=1}^{V} \exp\left({V_w}^{\top} V_{w_I}\right)}
\label{eq:4}
\end{equation}

\subsection{Offensive Expression Dataset}\label{sec:3.3}
In our study, offensive language refers to content that is insulting, disrespectful, abusive, or inappropriate, particularly in the context of online communication. We have annotated and classified the data into two categories: offensive and non-offensive. The non-offensive category may include neutral, informative, or general user comments that do not express hate, threats, or harmful intent. While we acknowledge the broader spectrum of toxic content categories such as hate speech, negative sentiment, or provocative statements, our work focuses specifically on a binary classification task (offensive vs. non-offensive) to establish a foundational model for Nastaliq Urdu content. Future work will aim to extend this framework by incorporating finer-grained categories (e.g., hate speech, sarcasm, profanity) for a more detailed offensive language taxonomy.

Urdu is a morphologically rich and context-sensitive language with several distinct characteristics that set it apart from many high-resource languages such as English. Firstly, Urdu is written in a variant of the Perso-Arabic script (Nastaliq), which is cursive, bidirectional, and more complex in visual structure than Latin scripts, posing unique challenges for NLP tasks. Secondly, Urdu exhibits high levels of inflection and agglutination, with words changing form based on gender, number, case, and politeness. Moreover, the language frequently incorporates words from Arabic, Persian, Hindi, and English, leading to lexical diversity and code-mixing. These linguistic features, along with limited NLP resources and annotated datasets, make offensive content detection in Urdu a particularly challenging and underexplored task.

Almost all available abusive content datasets for Urdu are constructed from the Twitter/X platform, where a 280-character limit constrains how users express opinions and introduces a systematic bias into the resulting corpora. We therefore build on a corpus collected from Facebook, where comments are not subject to such a limit. This work is the extension of our prior study (\C{hussain2022}{Hussain et al., 2022}) in which we collected comments from 19 Facebook Urdu pages in 2021. The details of selected Facebook pages is added in Table~\ref{tab:3}. The dataset indicates a good representative of various categories due to diversity of chosen pages. Some common categories include politics, religion. Following criteria is applied to select a particular page:

\begin{enumerate}
\item The minimum number of likes and followers to select a page is 30,000. This ensures active public Facebook pages to be selected.
\item The pages that use Urdu language mostly for opinion sharing.
\end{enumerate}

In (\C{hussain2022}{Hussain et al., 2022}), the collected comments were annotated by three annotators: two PhD students and one MS student, all native Urdu speakers with prior experience of data annotation. Inter-annotator agreement was measured using Fleiss' kappa [42], yielding a score of 0.88. Since the annotated pool was imbalanced, an equal number of instances was drawn at random from each class among the 12,416 preprocessed comments, producing a balanced set of 7,604 instances.

\begin{table}[htbp]
\centering
\caption{Chosen Facebook pages}
\label{tab:3}
\small
\begin{tabular}{@{}rlrr@{}}
\toprule
\multirow{2}{*}{\textbf{S\#}} & \multirow{2}{*}{\textbf{Facebook pages}} & \textbf{Before Pre-processing} & \textbf{After Pre-processing} \\
 & & \textbf{\# of Posts} & \textbf{\# of Posts} \\
\midrule
1 & Dunya News & 1000 & 460 \\
2 & Samaa TV & 300 & 167 \\
3 & Express News & 1200 & 780 \\
4 & Bol News & 450 & 120 \\
5 & ARY News & 800 & 333 \\
6 & GeoUrduNews & 8000 & 3000 \\
7 & Urdu Point & 2300 & 550 \\
8 & Daily Pakistan & 2800 & 880 \\
9 & BBC Urdu & 5000 & 900 \\
10 & ZemTv & 300 & 80 \\
11 & PTIOfficial & 7000 & 4400 \\
12 & PPPOfficial & 800 & 130 \\
13 & PML N Official & 1600 & 400 \\
14 & Pervaiz Hood & 50 & 10 \\
15 & JIPOfficial & 80 & 23 \\
16 & JUIPK & 300 & 78 \\
17 & Chingari & 400 & 80 \\
18 & marvisirmed & 20 & 5 \\
19 & Liberal2016 & 80 & 20 \\
\midrule
 & \textbf{Total} & \textbf{32,480} & \textbf{12,416} \\
\bottomrule
\end{tabular}
\end{table}

\textbf{Corpus curation in the present study:} Examination of this corpus revealed two categories of defect. The first is textual: missing letters within words, omitted tokens, and sentences left incomplete during collection. The second concerns the labels themselves, where a subset of comments had been assigned a class inconsistent with the annotation criteria. We therefore carried out a manual revision, correcting text-level inconsistencies and re-annotating every instance against the original definition of offensive content. The revision was performed by three annotators, all native Urdu speakers, none of whom had participated in the original annotation; re-annotation was thus independent of the earlier labels.

Inter-annotator agreement on the revised labels, again measured by Fleiss' kappa, improved to 0.90, and instances on which the three annotators disagreed were resolved by majority vote. The revision changed the class assigned to 500 comments, approximately 6.6\% of the corpus, with reassignments occurring in both directions. Because these reassignments did not offset exactly, class balance was restored by randomly removing instances from the larger class, yielding a final corpus of 3,802 offensive and 3,802 non-offensive instances. The curated corpus is publicly available (see Data Availability section), together with the annotation guidelines and a datasheet documenting its collection, preprocessing, and intended use.

\subsubsection{Dataset Cleaning and Pre-processing}\label{sec:3.3.1}
In the first step, the dataset is considered for screening, i.e. Garbage text was detected in the body of posts. Examples are empty text, images, URLs, hashtags, and timestamps, etc. So, we cleaned the dataset before processing it for further steps. We wrote a Python script to deal with this task. In the next step, we applied the following pre-processing steps:

\begin{itemize}
\item Removal of empty posts.
\item Removal of duplicate posts.
\item Fill in the missing characters/words in the sentences.
\item Removal of punctuations, mentions, hashtags, numbers, HTML tags, and URLs.
\item Replacement of Emojis with their corresponding text.
\item Manual translation of some English words into Urdu.
\item Removal of Urdu Stop words (not for transformer models)
\item Unicode normalization
\item Removal of zero-width characters
\item Normalization of common character variants (such as Yeh and Heh)
\item Removal of diacritics.
\end{itemize}

Some of the pre-processing steps are demonstrated in Table~\ref{tab:4}, with Urdu text added in the left column and the corresponding step is applied in the right column with English translation on both sides.

\begin{table}[htbp]
\centering
\caption{Demonstration of pre-processing steps on Urdu corpus with English translation}
\label{tab:4}
\small
\renewcommand{\arraystretch}{1.1}
\begin{tabular}{|>{\raggedright\arraybackslash}p{0.465\linewidth}|>{\raggedright\arraybackslash}p{0.465\linewidth}|}
\hline
\multicolumn{2}{|c|}{\textbf{Urdu Text Preprocessing}} \\ \hline
\multicolumn{2}{|c|}{\cellcolor{secbg}\textbf{1.\quad Punctuation Removal}} \\ \hline
\multicolumn{1}{|c|}{\cellcolor{subbg}\textbf{Before}} & \multicolumn{1}{c|}{\cellcolor{subbg}\textbf{After}} \\ \hline
\urb{t4a1}Quaid-e-Azam used to read speeches after watching them and he was also very handsome ------ & \urb{t4a2}Quaid-e-Azam used to read speeches after watching them and he was also very handsome \\ \hline
\multicolumn{2}{|c|}{\cellcolor{secbg}\textbf{2.\quad Stop word Removal}} \\ \hline
\multicolumn{1}{|c|}{\cellcolor{subbg}\textbf{Before}} & \multicolumn{1}{c|}{\cellcolor{subbg}\textbf{After}} \\ \hline
\urb{t4b1}Now the police will take bribes for this just like they do for drugs & \urb{t4b2}Police bribes drugs \\ \hline
\multicolumn{2}{|c|}{\cellcolor{secbg}\textbf{3.\quad Replace Emoji with Corresponding Text}} \\ \hline
\multicolumn{1}{|c|}{\cellcolor{subbg}\textbf{Before}} & \multicolumn{1}{c|}{\cellcolor{subbg}\textbf{After}} \\ \hline
\urb{t4c1}Everyone's sleep has become forbidden due to the fear of death, cowardly man \twemoji{1f602} & \urb{t4c2}Everyone's sleep has become forbidden due to the fear of death, cowardly man, tears of joy \\ \hline
\multicolumn{2}{|c|}{\cellcolor{secbg}\textbf{4.\quad Hashtag, Numbers and Link Removal}} \\ \hline
\multicolumn{1}{|c|}{\cellcolor{subbg}\textbf{Before}} & \multicolumn{1}{c|}{\cellcolor{subbg}\textbf{After}} \\ \hline
\urb{t4d1}A major European country has announced a billion-dollar package for Pakistan for the first time, \url{https://reut.r/3P4qHlE} & \urb{t4d2}A major European country has announced a billion-dollar package for Pakistan for the first time \\ \hline
\urb{t4e1}Pak-Sri Lanka match suddenly stopped \#PAKvsSL & \urb{t4e2}Pak-Sri Lanka match suddenly stopped \\ \hline
\urb{t4f1}I myself lived in Saudi Arabia 15 years ago, they did not understand it well. & \urb{t4f2}I myself lived in Saudi Arabia years ago, they did not understand it well. \\ \hline
\end{tabular}
\end{table}

After applying the pre-processing steps, the resultant dataset has 7604 instances. The dataset is balanced and the stratified sampling technique is applied to split the dataset into 80-20. The 80\% data is further divided into 90-10 in which 90\% is actually used for training and 10\% is used for validation purposes. The original 20\% is used for testing purposes. The number of instances for each part of the dataset is presented in Table~\ref{tab:5}.

\begin{table}[htbp]
\centering
\caption{Urdu OE corpus}
\label{tab:5}
\small
\begin{tabular}{@{}lcccccc@{}}
\toprule
\textbf{Dataset} & \textbf{Offensive} & \textbf{Not-Offensive} & \textbf{Total} & \textbf{Train} & \textbf{Validate} & \textbf{Test} \\
\midrule
Urdu & 3802 & 3802 & 7604 & 5475 & 608 & 1521 \\
\bottomrule
\end{tabular}
\end{table}

The Urdu stop words are removed from the dataset for only word2vec model that is used in the baselines. For language models, stop words are not removed. Some examples of Urdu stop words are presented in Table~\ref{tab:6}.

\begin{table}[htbp]
\centering
\caption{Sample list of Urdu stop words}
\label{tab:6}
\small
\renewcommand{\arraystretch}{1.5}
\begin{tabular}{@{}clclcl@{}}
\toprule
\textbf{Urdu} & \textbf{Translation} & \textbf{Urdu} & \textbf{Translation} & \textbf{Urdu} & \textbf{Translation} \\
\midrule
\ur{sw1} & Although & \ur{sw2} & Someone & \ur{sw3} & Was \\
\ur{sw4} & Seems & \ur{sw5} & Same & \ur{sw6} & Be \\
\ur{sw7} & Ever & \ur{sw8} & My & \ur{sw9} & Me \\
\ur{sw10} & I & \ur{sw11} & My & \ur{sw12} & No \\
\bottomrule
\end{tabular}
\end{table}

\subsection{Language Models}\label{sec:3.4}
\begin{figure}[!htbp]
\centering
\includegraphics[width=0.72\linewidth]{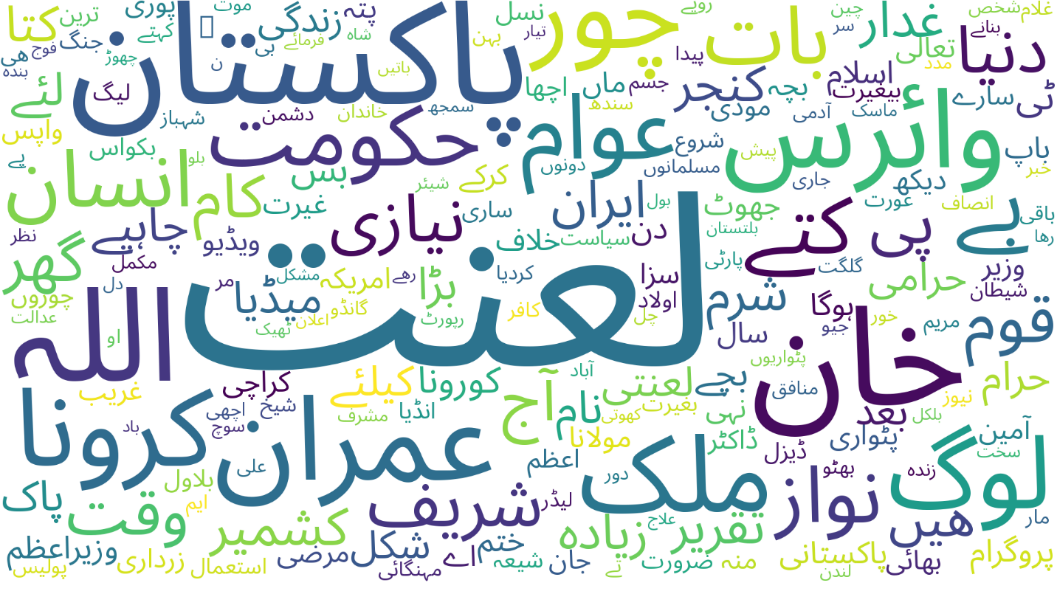}
\caption{Word cloud representation of the offensive dataset}
\label{fig:2}
\end{figure}

In this section, we described the details of four transformer models. These models are compared (in the result section) to define the best model for the design of a robust OED system (proposed framework) for the Nastaliq Urdu language. As described earlier, fine-tuning of these models with grid search hyper-parameters optimization are conducted. The models are Urdu-BERT, Urdu-RoBERTa, Urdu-Twhin-BERT, and MuRIL transformers. The word cloud representation of the dataset is presented in Fig.~\ref{fig:2}.

\paragraph{Urdu-BERT.}
BERT is a language model based on the transformer architecture (\C{devlin2018}{Devlin, Chang, Lee, \& Toutanova, 2018}) and it was developed by a team in 2018. It is a DL model in which each output component is connected to each input component and their weights are updated on the basis of connections. We are interested in exploring the potential of Urdu-BERT for the identification of OE in the Urdu language on the Facebook platform. The Urdu-BERT model has already demonstrated benchmark performance for various NLP tasks (\C{malik2023e}{M. S. I. Malik, Nazarova, Jamjoom, \& Ignatov, 2023}).

\paragraph{Urdu-Twhin-BERT.}
This is a recent Tweet Transformer model trained on 7 billion tweets. It is multilingual and trained on more than 100 languages and not only on text-based but also on ``rich social engagements'', i.e.Twin. So it differs from prior pre-trained transformer models. This model can be used in replacement to the BERT model for a variety of text mining and NLP tasks. We utilized the Urdu-Twhin-BERT transformer with a fine-tuning approach and compared its performance with Urdu-BERT, Urdu-RoBERTa, and MuRIL models.

\paragraph{Urdu-RoBERTa.}
The RoBERTa language model was developed after the BERT transformer in 2019 (\C{liu2019}{Y. Liu et al., 2019}). It uses the architecture of BERT as the basis but applies a diverse range of hyperparameters and removes the next sentence pre-training. This model presented an effective performance for various NLP and text-mining tasks (\C{rehan2023}{Rehan, Malik, \& Jamjoom, 2023}). We explored the Urdu-RoBERTa transformer with hyper-parameter fine-tuning for OED from Urdu tweets.

\paragraph{MuRIL.}
The MuRIL language model is one of the latest transformer models launched in the year 2020. The goal of designing this model is to support certain tasks in the regional languages of India, especially addressing spelling variations and transliteration tasks (\C{hu2020}{Hu et al., 2020}). The MuRIL model presented an effective performance in threatening and violent content identification (\C{khan2024}{Khan, Malik, \& Nadeem, 2024}; \C{rehan2023}{Rehan et al., 2023}). We are interested in exploring the strengths of MuRIL model to design an OED framework for Nastaliq Urdu content.

\subsubsection{Tokenization}\label{sec:3.4.1}
In this section, a description of tokenization and representation of data in appropriate format is provided. The tokenization of Facebook posts into a uniform format is performed in the first step. We have appended each Facebook post with the `CLS' token at the start and the `SEP' token at the end. By doing this, we can identify from where each sentence is started and this process results in a single vector for each Facebook post. The transformed vector is fed to the transformer model. For each transformer model, its specific tokenizer is used to tokenize the Facebook posts. For example, the following tokenizer (\C{zhang2022}{X. Zhang et al., 2022}) is used for Urdu-Twhin-BERT and following tokenizer (\url{https://huggingface.co/urduhack/roberta-urdu-small}) is used for Urdu-RoBERTa. After tokenizing, each post is now transformed into several tokens and tokens are mapped into indexes so that data can be fed to the language models.

\subsubsection{Fine-tuning}\label{sec:3.4.2}
In this section, the fine-tuning process for eight hyperparameters used to optimize the transformer models is presented. Two optimization strategies, namely manual tuning and grid search, are employed to identify the optimal hyperparameter values, following recommendations from prior studies (\C{malik2023e}{M. S. I. Malik, Nazarova, et al., 2023}; \C{malik2024c}{M. S. I. Malik, Younas, Jamjoom, \& Ignatov, 2024}). The explored ranges of these hyperparameters are provided in Table~\ref{tab:7}, while the optimal values are determined using the grid search approach. To evaluate the generalizability of the proposed framework, five widely used optimizers are investigated, including AdamW, Adam, AdaBelief, AdaBound, and AdaBoB (\C{xiang2025}{Xiang, Wang, Lei, \& Song, 2025}). The cross-entropy loss function is adopted for model training, while L2 regularization is applied as the regularization strategy.

No transformer layers are frozen during fine-tuning, and a linear decay schedule is employed for post-warmup learning rate adjustment. In addition, early stopping is not applied. To assess the impact of input length on classification performance, two sequence lengths (64 and 128) are explored, with the highest performance achieved using a sequence length of 128. Similarly, batch sizes of 16, 32, and 64 are evaluated to identify the most effective configuration. Four learning rate values are also examined to analyze their influence on model convergence and classification accuracy. The remaining hyperparameters are systematically explored to determine their optimal settings for fine-tuning the four transformer models. Furthermore, the step sizes for four hyperparameters (weight decay, warmup ratio, hidden dropout, and attention dropout) are presented in Table~\ref{tab:7}.

The stratified data splitting model is applied to divide the OE dataset into train, validate, and test parts. Initially, an 80-20 split is applied where 20\% is used for testing and 80\% is further divided into 90-10. The 90\% data is used for training purposes and 10\% is used for validation purposes. The details of the four language models are already provided at the start of section~\ref{sec:3.4}. We have used base versions of all transformers (Urdu-BERT, Urdu-RoBERTa, Urdu-Twhin-BERT, and MuRIL transformers). In addition, the pre-trained data includes Nastaliq Urdu language for the chosen Transformers. Each transformer model consists of a hidden\_size of 768 hidden\_layers, and attention-heads are 12. Each transformer model is trained, validated, and tested on each epoch and the performance on training and validation data is calculated using training and validation loss, accuracy, and macro F1-score. We utilized the HPC university cloud and Google Colab platforms to complete the fine-tuning experiments.

\paragraph{Regularization Pipeline.}
To improve generalization and reduce overfitting during fine-tuning, a complete regularization pipeline is employed. This includes hidden dropout and attention dropout tuning, L2 regularization through weight decay, warmup-based learning rate scheduling, linear learning-rate decay after warmup, conservative learning-rate exploration, and controlled epoch selection. No transformer layers are frozen during fine-tuning, allowing full parameter adaptation, while early stopping is intentionally not applied to maintain consistent comparison across transformer models. These settings collectively ensure stable optimization and effective generalization of the proposed framework.

\begin{table}[htbp]
\centering
\caption{Range of values for eight hyper-parameters (\C{malik2023e}{M. S. I. Malik, Nazarova, et al., 2023}; \C{malik2024c}{M. S. I. Malik, Younas, et al., 2024})}
\label{tab:7}
\small
\begin{tabular}{@{}ll@{}}
\toprule
\textbf{Hyperparameters} & \textbf{Grid Search} \\
\midrule
Sequence length & 64, 128 \\
Batch size & 16, 32, 64 \\
Learning rate & 3e-5, 2e-5, 1e-5, 1.5e-5, 0.99e-5 \\
Weight decay & 0.01--0.1 (step size = 0.01) \\
Warmup ratio & 0.06--0.1 (step size = 0.001) \\
Hidden dropout & 0.005--0.1 (step size = 0.01) \\
Attention dropout & 0.005--0.1 (step size = 0.01) \\
Epochs & 1--15 \\
\bottomrule
\end{tabular}
\end{table}

\subsubsection{Classification}\label{sec:3.4.3}
For the classification of offensive content, a layer is added on the top of four language models. This layer applies the softmax function and categorizes the content into offensive or non-offensive. The processed Facebook posts are passed to the softmax function and the training of language models is performed with the goal of optimizing the cross-entropy loss.

\subsubsection{Catastrophic Forgetting}\label{sec:3.4.4}
When we fine-tune a transformer model to learn new knowledge, it can encounter the problem of ``catastrophic forgetting'', that is forgetting the prior knowledge (\C{malik2023a}{M. S. I. Malik, 2023}; \C{malik2023b}{M. S. I. Malik, Cheema, \& Ignatov, 2023}). This is a common problem in transfer learning and we handle this by exploring a range of learning rates' values. Our analysis concluded that higher learning rates result in failure and poor convergence whereas low learning rates produce promising performance such as learning rates $\leq 10^{-5}$ delivered the highest accuracy and F1-score.

\subsubsection{Overfitting and under-fitting}\label{sec:3.4.5}
Overfitting and underfitting are common challenges in deep learning model training and directly affect generalization performance. To analyze the fitting behavior of the proposed framework, training and validation loss, accuracy, and macro F1-score are monitored across epochs during fine-tuning. Different epoch configurations (5-15) are explored to identify stable convergence behavior and appropriate fitting. The final epoch selection is based on validation performance and convergence stability. The observed training dynamics indicate that the selected configuration provides effective learning without substantial overfitting or underfitting, as further demonstrated in Section~\ref{sec:5.1}.

\subsection{Interpretability}\label{sec:3.5}
To make the classification system transparent and trustworthy, we use the SHAP XAI method (\C{liu2024}{J. Liu et al., 2024}) to explain the predictions of the proposed framework. The SHAP is a well known post-hoc XAI model, demonstrated its effectiveness for various classification tasks. In this study, we used SHAP to explain the prediction on the word level for individual instances. Furthermore, since Urdu is a right-to-left language and SHAP does not natively support its text alignment, this issue was manually addressed for proper visualization and interpretation.

\section{Experimental Setup}\label{sec:4}
In the previous section, we presented the description of the dataset, detail of transformers, and fine-tuning process. This section presents the detail of comparable models and performance metrics.

\subsection{Comparable Models}\label{sec:4.1}
To establish the effectiveness of the proposed framework, it needs to be compared with state-of-the-art models. For this purpose, we designed twelve comparable models using word2vec (\C{malik2023d}{M. S. I. Malik, Nawaz, Jamjoom, \& Ignatov, 2023}), RoBERTa, and MuRIL as the feature models and LR (\C{abbas2023}{Abbas \& Malik, 2023}), SVM (\C{ali2023a}{G. Ali \& Malik, 2023}; \C{malik2017}{M. Malik \& Hussain, 2017}), RF (\C{malik2023c}{M. S. I. Malik, Imran, \& Mamdouh, 2023}; \C{mehboob2021}{Mehboob \& Malik, 2021}), and CNN (\C{malik2020b}{M. S. I. Malik, 2020}) as the classifiers. These models demonstrated effective performance in various NLP tasks (\C{malik2024a}{M. S. I. Malik \& Nawaz, 2024}; \C{malik2023e}{M. S. I. Malik, Nazarova, et al., 2023}; \C{malik2024b}{M. S. I. Malik, Rehman, \& Ignatov, 2024}; \C{malik2024c}{M. S. I. Malik, Younas, et al., 2024}; \C{nawaz2022}{Nawaz \& Malik, 2022}). The details of the baseline models are provided below:

\begin{itemize}
\item Word2vec (100) + LR
\item Word2vec (100) + SVM
\item Word2vec (100) + RF
\item Word2vec (100) + CNN
\item RoBERTa + LR
\item RoBERTa + SVM
\item RoBERTa + RF
\item RoBERTa + CNN
\item MuRIL + LR
\item MuRIL + SVM
\item MuRIL + RF
\item MuRIL + CNN
\end{itemize}

The reasoning for choosing particular feature extraction method and ML model is provided below:

\textbf{Word2vec:} It is a powerful embedding technique that captures semantic relationships between words by mapping them into continuous vector space based on their context. It effectively preserves syntactic and semantic similarities, making it highly useful for downstream NLP tasks such as text classification, clustering, and sentiment analysis. For word2vec, a skip-gram model and 100 dimensions are configured.

The description and strengths of RoBERTa and MuRIL models are already presented in section~\ref{sec:3.4}.

\textbf{Logistic Regression:} It is a simple yet effective classification algorithm that models the probability of a binary outcome using a linear decision boundary. It is computationally efficient, interpretable, and performs well on linearly separable data.

\textbf{Support Vector Machine:} This is a robust and powerful classification algorithm that excels at handling high-dimensional data and works well with both linear and non-linear decision boundaries using kernel functions. It is less prone to overfitting and particularly strong in maximizing the margin between classes, leading to better generalization.

\textbf{Random Forest:} It is a versatile and powerful ensemble learning algorithm that combines multiple decision trees to improve classification accuracy and reduce overfitting. It handles both categorical and numerical data well and is robust to noise and outliers.

\textbf{Convolutional Neural Network:} They are highly effective at capturing local patterns and hierarchical features. They excel in extracting spatial and contextual relationships through convolutional layers. In NLP, CNNs are particularly useful for tasks like sentence classification and sentiment analysis by identifying key n-gram features.

We have added the details of parameters with their corresponding values for the baseline models in Table~\ref{tab:8}.

\subsection{Evaluation Metrics}\label{sec:4.2}
To evaluate the performance of classifiers, four standard metrics (\C{malik2018}{M. Malik \& Hussain, 2018}, \C{malik2020a}{2020}) are chosen, i.e. recall, precision, F1-score, and accuracy. As the test partition contains a fixed number of instances, point estimates of accuracy carry sampling uncertainty that aggregate figures alone do not convey. We therefore report 95\% Wilson score intervals on test accuracy alongside each point estimate. The Wilson interval is preferred to the normal approximation for proportions near the boundaries of the unit interval and does not require the interval to be symmetric about the estimate. All intervals are computed on the common test partition of 1,521 instances described in Section~\ref{sec:3.3}. In addition, two MARS measures are used to identify the unique behavior of each classifier. The measures are MARS Shine-Through and MARS Occlusion (\C{restrepo2022}{Restrepo, Mali, Abrahams, \& Ractham, 2022}). The detail of both measures is presented below.

Recently, two metrics are introduced to specify how many unique instances of the positive class are correctly classified by a particular classifier that other classifiers missed in comparison. Some recent studies utilized these state-of-the-art measures to verify the uniqueness of their proposed frameworks (\C{younas2023}{Younas, Malik, \& Ignatov, 2023}). The MARS Shine-Through measure is defined as the ratio of ``unique positive samples correctly predicted by a particular classifier'' to the ``unique positive samples correctly predicted by all the classifiers under evaluation''. This measure can be used to evaluate the performance of individual or combined classifiers. The second metric, MARS Occlusion, is defined as the ratio of ``false negative samples predicted by a particular classifier'' to the ``total unique positive samples correctly predicted by all the classifiers under evaluation''. So the denominator is the union of all classifiers' unique correctly predicted positive samples. The pseudocode of MARS measures is available publicly at \url{https://github.com/SoftwareImpacts/SIMPAC-2021-191}.

All the experiments are conducted using HuggingFace Transformers 4.20, SHAP 0.40, and other standard Python libraries. Model training is performed on Google Colab Pro (T4 GPU with 16GB RAM) as well as the university's HPC infrastructure equipped with an A100 GPU (40GB RAM).

\begin{table}[htbp]
\centering
\caption{Tuning parameters for the baseline models}
\label{tab:8}
\small
\begin{tabular}{@{}lll@{}}
\toprule
\textbf{Baseline Models} & \textbf{Tuning Parameter} & \textbf{Value} \\
\midrule
\multirow{3}{*}{Word2vec} & Architecture & Skip-gram \\
 & Window Size & 10 \\
 & Embedding Size & 100 \\
\midrule
\multirow{2}{*}{RoBERTa} & Max Sequence Length & 128 \\
 & Output\_hidden\_states & 12 (Base) \\
\midrule
\multirow{2}{*}{MuRIL} & Max Sequence Length & 128 \\
 & Output\_hidden\_states & 12 (Base) \\
\midrule
\multirow{5}{*}{Logistic Regression} & Solver & Liblinear \\
 & Maximum Iteration & 2000 \\
 & Class Weight & Balanced \\
 & C & 0.1 \\
 & Penalty & L2 (Ridge) \\
\midrule
\multirow{3}{*}{Support Vector Machine} & Kernel & Radial Basis Function \\
 & Regularization Parameter (C) & 10 \\
 & Gamma & 0.01 \\
\midrule
\multirow{5}{*}{Random Forest} & Number of Trees & 100 \\
 & Max Depth & Default \\
 & Min Samples per Split & 5 \\
 & Min Samples per Leaf & 2 \\
 & Max Features & Sqrt (n\_features) \\
\midrule
\multirow{5}{*}{Convolutional Neural Network} & Dropout Rate & 0.2, 0.3 \\
 & Learning Rate & 1e-3, 1e-4, 1e-5 \\
 & Optimizer & Adam \\
 & Batch Size & 32 \\
 & Filters & 64, 128, 256 \\
\bottomrule
\end{tabular}
\end{table}

\section{Results and Analysis}\label{sec:5}
This section describes the experiments and results conducted to evaluate the effectiveness of the proposed framework and its comparison with baselines. Furthermore, detail of SHAP XAI experiments is presented.

\subsection{Fine-tuning Phase}\label{sec:5.1}
This section presents the fine-tuning of four transformer models using the grid search hyperparameter optimization technique, with the Nastaliq Urdu corpus employed as part of the experimental setup. The primary objective is to identify the best-performing transformer model among the four selected candidates, while the explored hyperparameter configurations have already been described in Table~\ref{tab:7}. Each language model is trained, validated, and tested using the respective dataset partitions.

The grid search approach is applied to determine the optimal values of eight hyperparameters for each language model, and the top four performing configurations for each transformer are reported in Table~\ref{tab:9}. The fine-tuning procedure described in Section~\ref{sec:3.4.2} is consistently followed for all transformer models, and the results are presented in ascending order to improve clarity and reader comprehension.

To assess the impact of input representation length, sequence lengths of 64 and 128 are explored, with superior performance achieved using a sequence length of 128. The corresponding values of the selected hyperparameters are also reported alongside the performance metrics. For F1-score evaluation, class-wise performance is computed using both macro and weighted averaging schemes.

The fine-tuning results indicate that the Urdu-BERT model achieved a notable macro F1-score of 82.16\% and an accuracy of 82.18\% using a learning rate of $10^{-5}$, a hidden dropout rate of 0.005, and a weight decay of 0.01. In contrast, a higher learning rate of $2\times10^{-5}$, combined with alternative hyperparameter settings (hidden dropout of 0.1 and weight decay of 0.1), resulted in comparatively lower performance.

Similarly, Urdu-RoBERTa achieved its best performance under the same hyperparameter configuration, yielding an accuracy of 85.34\% and a macro F1-score of 85.31\%, demonstrating a clear improvement over Urdu-BERT. The fine-tuning results for Urdu-Twhin-BERT are presented in the lower section of Table~\ref{tab:9}, where the relationship between hyperparameter settings and corresponding performance metrics can be observed. Consistent with the previous models, the optimal results were obtained with a learning rate of $10^{-5}$, hidden dropout of 0.005, and weight decay of 0.01. Under this configuration, Urdu-Twhin-BERT achieved its best performance, attaining an accuracy of 86.91\% and a macro F1-score of 86.92\%.

Finally, the fine-tuning results of the MuRIL model are compared with those of the other three transformer models. MuRIL achieved the highest accuracy and macro F1-score among the four, attaining an accuracy of 87.05\% and a macro F1-score of 87.06\%, with a margin of 0.14 percentage points over Twhin-BERT. The optimal performance was achieved with a hidden dropout rate of 0.01 and a weight decay of 0.01. This concludes the fine-tuning and comparative evaluation of the four transformer models.

\begin{table}[htbp]
\centering
\caption{Performance of fine-tuning for BERT, RoBERTa, Twhin-BERT, and MuRIL on the test dataset}
\label{tab:9}
\small
\begin{threeparttable}
\begin{tabular}{@{}lccccccccc@{}}
\toprule
\multirow{2}{*}{\textbf{Models}} & \multirow{2}{*}{\textbf{B.S}} & \multirow{2}{*}{\textbf{L.R}} & \multirow{2}{*}{\textbf{H.D}} & \multirow{2}{*}{\textbf{W.D}} & \multirow{2}{*}{\textbf{Accuracy}} & \multicolumn{4}{c}{\textbf{F1-score}} \\
\cmidrule(l){7-10}
 &  &  &  &  &  & \textbf{OE} & \textbf{Not} & \textbf{Macro} & \textbf{Weighted} \\
\midrule
\multirow{4}{*}{Urdu-BERT} & 32 & 2e-5 & 0.1 & 0.1 & 80.87 & 80.19 & 81.50 & 80.85 & 80.86 \\
 & 32 & 1e-5 & 0.05 & 0.01 & 80.93 & 80.33 & 81.51 & 80.92 & 80.91 \\
 & 32 & 1e-5 & 0.1 & 0.01 & 81.13 & 81.75 & 80.46 & 81.11 & 81.12 \\
 & 32 & 1e-5 & 0.005 & 0.01 & \textbf{\underline{82.18}} & \textbf{\underline{81.55}} & \textbf{\underline{82.77}} & \textbf{\underline{82.16}} & \textbf{\underline{82.17}} \\
\midrule
\multirow{4}{*}{Urdu-RoBERTa} & 32 & 1e-5 & 0.1 & 0.01 & 83.74 & 83.42 & 84.06 & 83.73 & 83.74 \\
 & 32 & 1e-5 & 0.05 & 0.01 & 84.68 & 84.14 & 85.19 & 84.66 & 84.67 \\
 & 32 & 1e-5 & 0.01 & 0.01 & 85.21 & 84.77 & 85.62 & 85.19 & 85.20 \\
 & 32 & 1e-5 & 0.005 & 0.01 & \textbf{\underline{85.34}} & \textbf{\underline{84.65}} & \textbf{\underline{85.97}} & \textbf{\underline{85.31}} & \textbf{\underline{85.31}} \\
\midrule
\multirow{4}{*}{Urdu-Twhin-BERT} & 32 & 1e-5 & 0.01 & 0.01 & 85.86 & 85.40 & 86.30 & 85.85 & 85.86 \\
 & 32 & 1e-5 & 0.005 & 0.01 & 86.52 & 86.01 & 87.00 & 86.50 & 86.51 \\
 & 64 & 1e-5 & 0.005 & 0.05 & 86.72 & 86.44 & 86.98 & 86.71 & 86.72 \\
 & 64 & 1e-5 & 0.005 & 0.01 & \textbf{\underline{86.91}} & \textbf{\underline{86.90}} & \textbf{\underline{86.93}} & \textbf{\underline{86.92}} & \textbf{\underline{86.93}} \\
\midrule
\multirow{4}{*}{MuRIL} & 32 & 1e-5 & 0.005 & 0.1 & 86.19 & 86.52 & 85.85 & 86.20 & 86.19 \\
 & 32 & 1.5e-5 & 0.005 & 0.01 & 86.85 & 86.67 & 87.03 & 86.85 & 86.86 \\
 & 32 & 1e-5 & 0.1 & 0.1 & 86.07 & 85.79 & 86.32 & 86.06 & 86.06 \\
 & 32 & 1e-5 & 0.01 & 0.01 & \textbf{\underline{87.05}} & \textbf{\underline{87.23}} & \textbf{\underline{86.86}} & \textbf{\underline{87.06}} & \textbf{\underline{87.06}} \\
\bottomrule
\end{tabular}
\begin{tablenotes}[flushleft]\footnotesize
\item B.S: batch size; L.R: learning rate; H.D: hidden dropout; W.D: weight decay.
\end{tablenotes}
\end{threeparttable}
\end{table}

Next, the generality of the best-performing fine-tuned MuRIL model is further evaluated by replacing the Adam optimizer with four state-of-the-art optimizers, namely AdamW, AdaBelief, AdaBound, and AdaBoB. This is accomplished by keeping the remaining experimental settings unchanged, including batch size, sequence length, and training epochs. The objective of this analysis is to examine whether the proposed framework (fine-tuned MuRIL) maintains consistent performance across different optimizers, as one component of the stability assessment. The performance of the fine-tuned MuRIL model is presented in Table~\ref{tab:10}, under multiple hyperparameter settings for each optimizer, along with the corresponding accuracy and F1-score values.

The experimental results demonstrate a highly consistent performance trend across all five optimizers, indicating the stability of the proposed framework. For AdaBelief, the highest performance is achieved with a learning rate of $10^{-5}$, hidden dropout of 0.01, and weight decay of 0.01, resulting in an accuracy of 87.03\% and a macro F1-score of 87.05\%. Similarly, AdaBound, AdaBoB, and AdamW achieve nearly identical performance, each attaining a macro F1-score of 87.05\% with only marginal variations in accuracy. Adam records the highest values at 87.05\% accuracy and 87.06\% macro F1-score, though the spread across all five optimizers is 0.02 percentage points, corresponding to approximately 0.3 instances of the 1,521-instance test partition and therefore below the resolution of this evaluation, which aligns with the best fine-tuning results reported earlier for MuRIL.

This indicates that the fine-tuned MuRIL configuration is not sensitive to the choice of optimizer within the range examined. The observed differences are, however, too small for the test partition to resolve, so the result should be read as an absence of detectable optimizer dependence rather than as positive evidence of generalization, which would require evaluation on held-out data from a different source.

\begin{table}[htbp]
\centering
\caption{Generality evaluation of fine-tuned MuRIL model across five state-of-the-art optimizers}
\label{tab:10}
\small
\begin{threeparttable}
\begin{tabular}{@{}lcccccccc@{}}
\toprule
\multirow{2}{*}{\textbf{Optimizer}} & \multirow{2}{*}{\textbf{Final L.R}} & \multirow{2}{*}{\textbf{H.D}} & \multirow{2}{*}{\textbf{W.D}} & \multirow{2}{*}{\textbf{Accuracy}} & \multicolumn{4}{c}{\textbf{F1-score}} \\
\cmidrule(l){6-9}
 &  &  &  &  & \textbf{OE} & \textbf{Not} & \textbf{Macro} & \textbf{Weighted} \\
\midrule
\multirow{4}{*}{AdaBelief} & 1e-5 & 0.005 & 0.1 & 86.19 & 86.52 & 85.89 & 86.20 & 86.20 \\
 & 1.5e-5 & 0.05 & 0.01 & 86.88 & 86.68 & 87.02 & 86.89 & 86.88 \\
 & 2e-5 & 0.01 & 0.01 & 86.94 & 87.09 & 86.70 & 86.94 & 86.95 \\
 & 1e-5 & 0.01 & 0.01 & \textbf{\underline{87.03}} & \textbf{\underline{87.21}} & \textbf{\underline{86.86}} & \textbf{\underline{87.05}} & \textbf{\underline{87.05}} \\
\midrule
\multirow{4}{*}{AdaBound} & 1e-5 & 0.005 & 0.1 & 86.20 & 86.53 & 85.86 & 86.19 & 86.20 \\
 & 1.5e-5 & 0.05 & 0.01 & 86.89 & 86.69 & 86.95 & 86.89 & 86.88 \\
 & 2e-5 & 0.01 & 0.01 & 86.95 & 87.09 & 86.70 & 86.95 & 86.96 \\
 & 1e-5 & 0.01 & 0.01 & \textbf{\underline{87.03}} & \textbf{\underline{87.22}} & \textbf{\underline{86.86}} & \textbf{\underline{87.05}} & \textbf{\underline{87.05}} \\
\midrule
\multirow{4}{*}{AdaBoB} & 1e-5 & 0.005 & 0.1 & 86.18 & 86.51 & 85.90 & 86.19 & 86.19 \\
 & 1.5e-5 & 0.05 & 0.01 & 86.89 & 86.69 & 87.01 & 86.88 & 86.88 \\
 & 2e-5 & 0.01 & 0.01 & 86.95 & 87.09 & 86.69 & 86.95 & 86.95 \\
 & 1e-5 & 0.01 & 0.01 & \textbf{\underline{87.04}} & \textbf{\underline{87.21}} & \textbf{\underline{86.86}} & \textbf{\underline{87.05}} & \textbf{\underline{87.05}} \\
\midrule
\multirow{4}{*}{(AdamW)} & 1e-5 & 0.005 & 0.1 & 86.20 & 86.53 & 85.86 & 86.20 & 86.20 \\
 & 1.5e-5 & 0.005 & 0.01 & 86.87 & 86.67 & 87.03 & 86.89 & 86.89 \\
 & 2e-5 & 0.01 & 0.01 & 86.94 & 87.10 & 86.68 & 86.93 & 86.94 \\
 & 1e-5 & 0.01 & 0.01 & \textbf{\underline{87.03}} & \textbf{\underline{87.22}} & \textbf{\underline{86.87}} & \textbf{\underline{87.05}} & \textbf{\underline{87.05}} \\
\midrule
\multirow{4}{*}{(Adam)} & 1e-5 & 0.005 & 0.1 & 86.19 & 86.52 & 85.85 & 86.20 & 86.19 \\
 & 1.5e-5 & 0.005 & 0.01 & 86.85 & 86.67 & 87.03 & 86.85 & 86.86 \\
 & 1e-5 & 0.1 & 0.1 & 86.07 & 85.79 & 86.32 & 86.06 & 86.06 \\
 & 1e-5 & 0.01 & 0.01 & \textbf{\underline{87.05}} & \textbf{\underline{87.23}} & \textbf{\underline{86.86}} & \textbf{\underline{87.06}} & \textbf{\underline{87.06}} \\
\bottomrule
\end{tabular}
\begin{tablenotes}[flushleft]\footnotesize
\item Batch size = 32, Sequence length = 128, Epochs = 15. Other parameter values: $\beta_1 = 0.9$, $\beta_2 = 0.999$, Epsilon = 1e-8.
\item Parameters valid for AdaBelief, AdaBound, and AdaBoB ($\alpha_f = 0.1$, $\gamma = 10^{-3}$).
\end{tablenotes}
\end{threeparttable}
\end{table}

\begin{figure}[!htbp]
\centering
\includegraphics[page=1,width=0.75\linewidth]{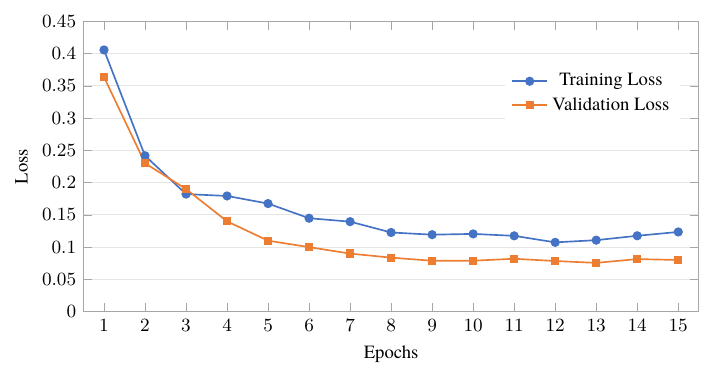}
\caption{Response of Loss on train, and validate parts by fine-tuned MuRIL [for 15 epochs]}
\label{fig:3}
\end{figure}

To quantitatively analyze the convergence behavior and fitting state of the proposed framework, the training and validation losses of the best-performing fine-tuned MuRIL model are presented in Fig.~\ref{fig:3}.

The results show a consistent decrease in both training and validation loss during the initial epochs, followed by stabilization at epoch 10 and onwards. The validation curve closely follows the training curve, indicating effective regularization and stable convergence behavior. Furthermore, the absence of a significant performance gap between training and validation metrics suggests that the model generalizes well and does not suffer from substantial overfitting or underfitting. This analysis further confirms the robustness and reliability of the proposed framework.

\subsection{Comparison with Baselines}\label{sec:5.2}
In this section, the performance of MuRIL model is compared with baseline models including three transformer models to analyze the effectiveness of the proposed framework, and a conclusion is drawn. Four evaluation metrics are used to evaluate the significance of classifiers and the results are presented in Table~\ref{tab:11}. The performance is presented in ascending order and for the F1-score, macro and weighted average are calculated. Considering baselines, MuRIL + SVM presented better performance than others by delivering 81.45\% macro F1-score and 81.53\% accuracy. In contrast, the four language models (Urdu-BERT, Urdu-RoBERTa, Urdu-Twhin-BERT, MuRIL) with fine-tuning demonstrated better performance than the twelve baselines in all performance metrics. Furthermore, the fine-tuned MuRIL model achieved the highest scores across all four metrics. The Urdu-Twhin-BERT model presented the second-best performance by showing 86.92\% macro F1-score and 86.91\% accuracy. In contrast, the MuRIL model after fine-tuning presented 87.05\% accuracy and 87.06\% macro F1-score. Relative to the strongest baseline (MuRIL + SVM), the fine-tuned MuRIL model improves accuracy by 5.52 percentage points, precision by 5.01, and recall by 5.52. The margin over the second-best transformer, Twhin-BERT, is 0.14 percentage points, corresponding to approximately two instances of the 1,521-instance test set and therefore within the resolution of this evaluation. The two models are best regarded as comparable, with MuRIL preferred on the secondary measures reported below.

\begin{table}[htbp]
\centering
\caption{Comparison of four fine-tuned transformers with twelve baselines [95\% CI denotes the Wilson score interval on test accuracy, $n = 1{,}521$]}
\label{tab:11}
\small
\begin{tabular}{@{}lcccccc@{}}
\toprule
\multirow{2}{*}{\textbf{Classifiers}} & \multirow{2}{*}{\textbf{Accuracy}} & \multirow{2}{*}{\textbf{95\% CI}} & \multicolumn{2}{c}{\textbf{F1-score}} & \multirow{2}{*}{\textbf{Precision}} & \multirow{2}{*}{\textbf{Recall}} \\
\cmidrule(lr){4-5}
 & & & \textbf{Macro} & \textbf{Weighted} & & \\
\midrule
Word2vec+SVM & 74.10 & [71.84, 76.24] & 73.75 & 73.75 & 75.46 & 74.10 \\
Word2vec+LR & 71.27 & [68.94, 73.49] & 70.96 & 70.95 & 72.24 & 71.28 \\
Word2vec+RF & 72.58 & [70.28, 74.76] & 72.43 & 72.43 & 73.09 & 72.59 \\
Word2vec+CNN & 73.31 & [71.03, 75.47] & 73.08 & 73.08 & 74.12 & 73.31 \\
Urdu-RoBERTa + SVM & 75.48 & [73.26, 77.58] & 75.48 & 75.48 & 75.48 & 75.48 \\
Urdu-RoBERTa + Logistic Reg & 78.04 & [75.89, 80.05] & 78.04 & 78.04 & 78.04 & 78.04 \\
Urdu-RoBERTa + Random Forest & 76.33 & [74.13, 78.40] & 76.27 & 76.27 & 76.60 & 76.33 \\
Urdu-RoBERTa + CNN & 77.84 & [75.68, 79.86] & 77.66 & 77.66 & 78.78 & 77.85 \\
MuRIL + SVM & \textbf{\underline{81.53}} & [79.50, 83.40] & \textbf{\underline{81.45}} & \textbf{\underline{81.45}} & \textbf{\underline{82.07}} & \textbf{\underline{81.53}} \\
MuRIL + Logistic Reg & 79.03 & [76.91, 81.00] & 78.72 & 78.72 & 80.81 & 79.03 \\
MuRIL + Random Forest & 80.60 & [78.54, 82.51] & 80.58 & 80.58 & 80.79 & 80.61 \\
MuRIL + CNN & 79.49 & [77.39, 81.44] & 79.35 & 79.35 & 80.32 & 79.49 \\
\midrule
Fine-tuned Urdu-BERT & 82.18 & [80.18, 84.02] & 82.16 & 82.18 & 82.34 & 82.19 \\
Fine-tuned Urdu-RoBERTa & 85.34 & [83.47, 87.03] & 85.31 & 85.34 & 85.63 & 85.34 \\
Fine-tuned Urdu-Twhin-BERT & 86.91 & [85.12, 88.51] & 86.92 & 86.92 & 86.91 & 86.92 \\
\textbf{Fine-tuned MuRIL} & \textbf{\underline{87.05}} & [85.27, 88.64] & \textbf{\underline{87.06}} & \textbf{\underline{87.06}} & \textbf{\underline{87.08}} & \textbf{\underline{87.05}} \\
\bottomrule
\end{tabular}
\end{table}

\begin{figure}[!htbp]
\centering
\includegraphics[page=2,width=1.0\linewidth]{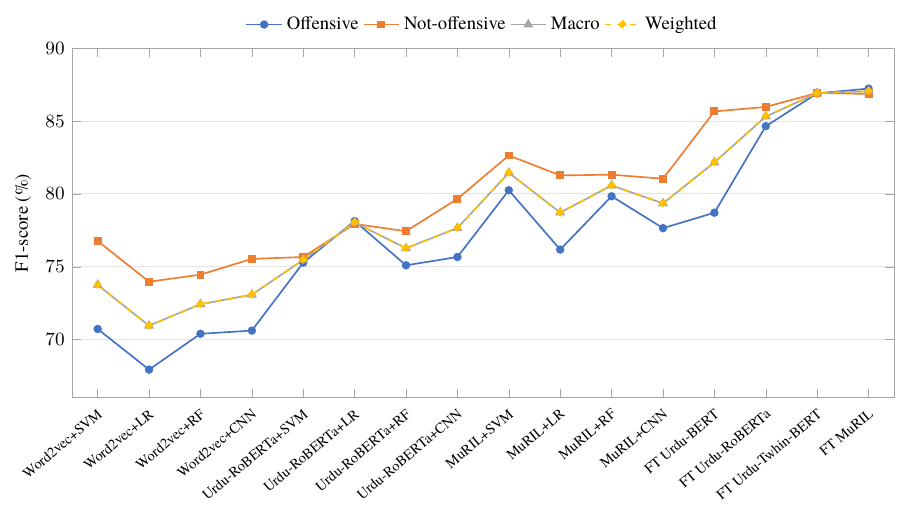}
\caption{Class-wise performance of classifiers and baselines (F1-score)}
\label{fig:4}
\end{figure}

Considering the class-wise performance of proposed and baseline models, we calculated the F1-score of each class, with macro and weighted average, and results are presented in Fig.~\ref{fig:4}. Dealing with offensive class detection, MuRIL model (with fine-tuning) outperformed all baseline models including three transformer models. The MuRIL+SVM model performed better than other baselines, and fine-tuned MuRIL improved the F1-score of offensive class detection by 6.98\%. Likewise, for non-offensive class identification, the MuRIL+SVM presented an 82.64\% F1-score which is less than the values obtained by the MuRIL and other transformers. Thus, an improvement of 4.22\% is observed for non-offensive class identification. For weighted average, again MuRIL (with fine-tuning) achieved the highest weighted-average F1-score at 87.06\%. This results in a 5.6\% improvement compared to baseline models. The evaluation was performed using four standard metrics across both classes.

Confidence intervals on test accuracy place these margins in context. On a test partition of 1,521 instances, 95\% Wilson intervals span approximately $\pm$1.7 percentage points for the fine-tuned transformers and $\pm$2.0 points for the baselines. The 5.52-point margin of the fine-tuned MuRIL model over the strongest baseline corresponds to 84 instances and exceeds the interval width by a factor of roughly three; this separation is therefore not attributable to sampling variation in the test partition. The 0.14-point margin over Twhin-BERT corresponds to approximately two instances and lies well inside the interval, as does the 1.71-point margin over Urdu-RoBERTa, at 26 instances. The comparison against the feature--classifier baselines is accordingly robust, whereas the ordering among the three strongest transformers is not resolved by a test set of this size. We note that these are marginal intervals on each estimate rather than a paired comparison; establishing significance for the differences between models would require a paired test such as McNemar's on the joint predictions, which we leave to future work.

In the end, experiments are performed to analyze the unique prediction demonstrated by each model compared to other models. The MARS Shine-Through and MARS Occlusion matrices are chosen to evaluate the strengths of proposed and baseline models in this regard. The first set of experiments is performed using the MARS Occlusion matrix. The MuRIL model (with fine-tuning) is compared with three transformers (with fine-tuning) and with four best baselines and the results are added in Table~\ref{tab:12}. MARS Occlusion identifies the ``ratio of false negative (wrongly predicted as non-offensive) instances predicted by a model to the total positive instances predicted by all other models''. For the combined model, the predictions are merged. It should be noted that these measures quantify the complementarity of predictions rather than overall predictive quality: they indicate which instances a model recovers or misses that others do not, and are therefore reported alongside, rather than in place of, the aggregate metrics above. The intersection of the x and y axes (each cell) presented two values, the left side value indicates the score predicted by a model on the y-axis and the right side value highlights the score achieved by the two models (on the intersection of the x-axis and y-axis). Considering the four transformer models, the MuRIL demonstrated the best performance by predicting only 9.8\% false negative instances compared to other models. The Urdu-Twhin-BERT model demonstrated 11.5\% whereas Urdu-RoBERTa produced 13.5\% and Urdu-BERT gave 19.7\%.

The four baseline models presented worse percentages compared to all transformers. Hence MuRIL recorded the lowest occlusion value among the standalone models. When combined models are considered, the MuRIL with RoBERTa+LR presented the lowest value at 3.9\% false negative instances. On the other hand, MuRIL combined with Urdu-RoBERTa delivered better performance when only fine-tuned models are considered. This indicates that the fine-tuned MuRIL model produces fewer false negatives than the alternatives considered.

\begin{table}[htbp]
\centering
\caption{Comparison of models using MARS Occlusion matrix}
\label{tab:12}
\scriptsize
\setlength{\tabcolsep}{4pt}
\begin{threeparttable}
\begin{tabular}{@{}clcccccccc@{}}
\toprule
 & \textbf{Models} & \multicolumn{8}{c}{\textbf{X-axis}} \\
\cmidrule(l){3-10}
 & & \textbf{R-LR} & \textbf{M-SVM} & \textbf{M-RF} & \textbf{M-CNN} & \textbf{U-BERT} & \textbf{U-RBRT} & \textbf{U-TBERT} & \textbf{MuRIL} \\
\midrule
\multirow{8}{*}{\rotatebox{90}{\textbf{Y-axis}}} & \textbf{R-LR} & -- & 0.201\,$|$\,0.098 & 0.201\,$|$\,0.088 & 0.201\,$|$\,0.083 & 0.201\,$|$\,0.086 & 0.201\,$|$\,0.059 & 0.201\,$|$\,0.048 & 0.201\,$|$\,0.039 \\
 & \textbf{M-SVM} & 0.235\,$|$\,0.098 & -- & 0.235\,$|$\,0.19 & 0.235\,$|$\,0.182 & 0.235\,$|$\,0.097 & 0.235\,$|$\,0.07 & 0.235\,$|$\,0.066 & 0.235\,$|$\,0.071 \\
 & \textbf{M-RF} & 0.217\,$|$\,0.088 & 0.217\,$|$\,0.19 & -- & 0.217\,$|$\,0.188 & 0.217\,$|$\,0.091 & 0.217\,$|$\,0.07 & 0.217\,$|$\,0.059 & 0.217\,$|$\,0.067 \\
 & \textbf{M-CNN} & 0.213\,$|$\,0.083 & 0.213\,$|$\,0.182 & 0.213\,$|$\,0.188 & -- & 0.213\,$|$\,0.076 & 0.213\,$|$\,0.064 & 0.213\,$|$\,0.05 & 0.213\,$|$\,0.058 \\
 & \textbf{U-BERT} & 0.197\,$|$\,0.086 & 0.197\,$|$\,0.097 & 0.197\,$|$\,0.091 & 0.197\,$|$\,0.076 & -- & 0.197\,$|$\,0.084 & 0.197\,$|$\,0.076 & 0.197\,$|$\,0.066 \\
 & \textbf{U-RBRT} & 0.135\,$|$\,0.059 & 0.135\,$|$\,0.07 & 0.135\,$|$\,0.07 & 0.135\,$|$\,0.064 & 0.135\,$|$\,0.084 & -- & 0.135\,$|$\,0.062 & 0.135\,$|$\,0.052 \\
 & \textbf{U-TBERT} & 0.115\,$|$\,0.048 & 0.115\,$|$\,0.066 & 0.115\,$|$\,0.059 & 0.115\,$|$\,0.05 & 0.115\,$|$\,0.076 & 0.115\,$|$\,0.062 & -- & 0.115\,$|$\,0.054 \\
 & \textbf{MuRIL} & \textbf{\underline{0.098\,$|$\,0.039}} & \textbf{\underline{0.098\,$|$\,0.071}} & \textbf{\underline{0.098\,$|$\,0.067}} & \textbf{\underline{0.098\,$|$\,0.058}} & \textbf{\underline{0.098\,$|$\,0.066}} & \textbf{\underline{0.098\,$|$\,0.052}} & \textbf{\underline{0.098\,$|$\,0.054}} & -- \\
\bottomrule
\end{tabular}
\begin{tablenotes}[flushleft]\footnotesize
\item R-LR: RoBERTa + LR; M-SVM: MuRIL + SVM; M-RF: MuRIL + RF; M-CNN: MuRIL + CNN; U-BERT: Urdu-BERT; U-RBRT: Urdu-RoBERTa; U-TBERT: Urdu-Twhin-BERT.
\end{tablenotes}
\end{threeparttable}
\end{table}

\begin{figure}[!htbp]
\centering
\includegraphics[width=0.75\linewidth]{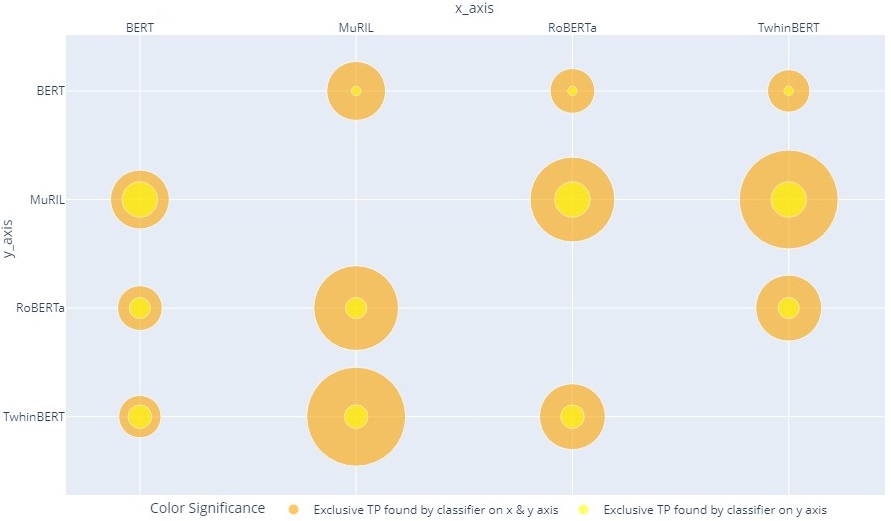}
\caption{MARS Shine-Through matrix}
\label{fig:5}
\end{figure}

Next, the MARS Shine-Through matrix is calculated considering only four transformer models to observe the unique offensive class instances identified by each model. The MARS Shine-Through identifies the ``ratio of exclusive OE instances predicted by a model to the total OE instances identified by all models''. The results are presented in Table~\ref{tab:13} and Fig.~\ref{fig:5}. It is evident that the MuRIL (with fine-tuning) demonstrated the highest value (2.1\%) for the prediction of unique OE instances which the other three transformers missed. The BERT model predicted only 1\% unique positive instances. Considering the combined model, MuRIL combined with Twhin-BERT presented 5.8\% unique positive class instances and exhibited the highest score. Furthermore, the visual description of the MARS Shine-Through matrix is presented in Fig.~\ref{fig:5}.

\begin{table}[htbp]
\centering
\caption{Fine-tuned transformers evaluation using MARS Shine-Through measure}
\label{tab:13}
\small
\setlength{\tabcolsep}{4pt}
\begin{threeparttable}
\begin{tabular}{@{}clcccc@{}}
\toprule
 & \textbf{} & \multicolumn{4}{c}{\textbf{X-axis}} \\
\cmidrule(l){3-6}
 & & \textbf{BERT} & \textbf{RoBERTa} & \textbf{Twhin-BERT} & \textbf{MuRIL} \\
\midrule
\multirow{4}{*}{\rotatebox{90}{\textbf{Y-axis}}} & \textbf{BERT} & -- & 0.01\,$|$\,0.026 & 0.01\,$|$\,0.025 & 0.01\,$|$\,0.034 \\
 & \textbf{RoBERTa} & 0.012\,$|$\,0.026 & -- & 0.012\,$|$\,0.039 & 0.012\,$|$\,0.05 \\
 & \textbf{Twhin-BERT} & 0.014\,$|$\,0.025 & 0.014\,$|$\,0.039 & -- & 0.014\,$|$\,0.058 \\
 & \textbf{MuRIL} & \textbf{\underline{0.021\,$|$\,0.034}} & \textbf{\underline{0.021\,$|$\,0.05}} & \textbf{\underline{0.021\,$|$\,0.058}} & -- \\
\bottomrule
\end{tabular}
\end{threeparttable}
\end{table}

The outer circles represent the performance of the combined models (e.g., BERT+MuRIL), while the inner circles illustrate the performance of individual models (e.g., BERT alone or Twhin-BERT alone). Among the inner circles, it is evident that the MuRIL (with fine-tuning) records the highest Shine-Through value, as indicated by its larger size. When MuRIL is combined with Twhin-BERT, the model attains the highest MARS Shine-Through score. This completes the evaluation of four transformer models using the MARS Shine-Through matrix.

\begin{figure}[!htbp]
\centering
\includegraphics[width=0.7\linewidth]{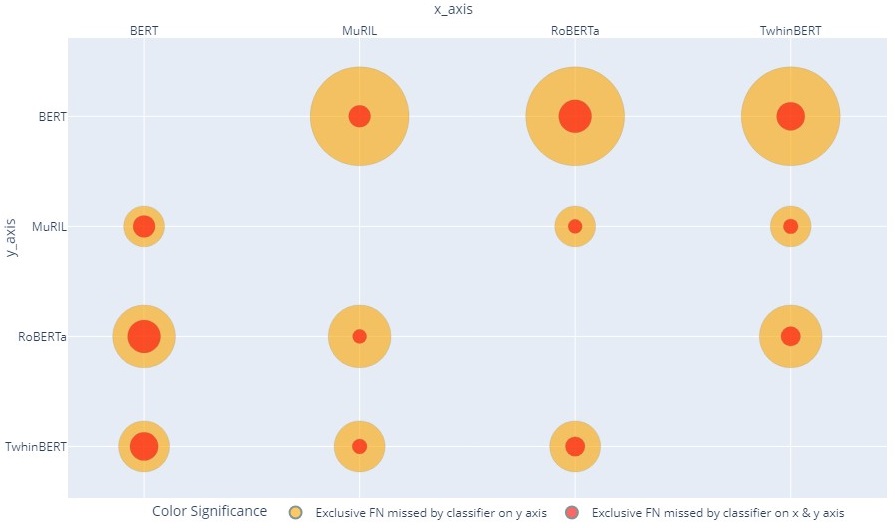}
\caption{MARS Occlusion matrix}
\label{fig:6}
\end{figure}

The last experiment is conducted to test the predictive capability of four transformer models using the MARS occlusion measure. The performance is evaluated on the basis of false negative instances predicted by an individual or combined model and shown in Table~\ref{tab:14} and Fig.~\ref{fig:6} respectively.

\begin{table}[htbp]
\centering
\caption{Fine-tuned transformers' evaluation using MARS occlusion measure}
\label{tab:14}
\small
\setlength{\tabcolsep}{4pt}
\begin{threeparttable}
\begin{tabular}{@{}clcccc@{}}
\toprule
 & \textbf{Transformers} & \multicolumn{4}{c}{\textbf{X-axis}} \\
\cmidrule(l){3-6}
 & & \textbf{BERT} & \textbf{RoBERTa} & \textbf{Twhin-BERT} & \textbf{MuRIL} \\
\midrule
\multirow{4}{*}{\rotatebox{90}{\textbf{Y-axis}}} & \textbf{BERT} & -- & 0.174\,$|$\,0.058 & 0.174\,$|$\,0.05 & 0.174\,$|$\,0.039 \\
 & \textbf{RoBERTa} & 0.11\,$|$\,0.058 & -- & 0.11\,$|$\,0.034 & 0.11\,$|$\,0.025 \\
 & \textbf{Twhin-BERT} & 0.09\,$|$\,0.05 & 0.09\,$|$\,0.034 & -- & 0.09\,$|$\,0.026 \\
 & \textbf{MuRIL} & \textbf{\underline{0.072\,$|$\,0.039}} & \textbf{\underline{0.072\,$|$\,0.025}} & \textbf{\underline{0.072\,$|$\,0.026}} & -- \\
\bottomrule
\end{tabular}
\end{threeparttable}
\end{table}

The minimum value is again recorded by the fine-tuned MuRIL model, i.e. 0.072, indicating that 7.2\% of offensive posts are misclassified as non-offensive when MuRIL is compared against the other transformers. Note that this value differs from the 0.098 reported in Table~\ref{tab:12} because the MARS Occlusion denominator depends on the comparison set: Table~\ref{tab:12} evaluates MuRIL against both transformers and baselines, whereas Table~\ref{tab:14} restricts the comparison to the four fine-tuned transformers. The two figures are therefore not directly comparable. In contrast, the BERT model presented the lowest performance by predicting 17\% false negative Facebook posts as an individual model.

As far as the combined models are considered, MuRIL plus RoBERTa performed best by presenting a value of 0.025, corresponding to 2.5\% of offensive posts being misclassified. A similar pattern is observed in Fig.~\ref{fig:6}, Note that the circle arrangement here is inverted relative to Fig.~\ref{fig:5}: the outer circles represent the performance of individual models, and the inner circles indicate the performance of combined models. Note also that the two measures run in opposite directions -- higher Shine-Through values are preferable in Fig.~\ref{fig:5}, whereas lower Occlusion values are preferable in Fig.~\ref{fig:6}. The BERT model demonstrates the lowest performance, reflected by the largest outer circle and relatively large inner circles when combined with other models. In contrast, the MuRIL model exhibits the best performance, with the smallest outer circle as an individual model and much smaller inner circles when used in combination. This concludes the comparison of MuRIL with three transformer models and with baseline models, on both MARS measures.

\subsection{Word-Level Attribution Analysis}\label{sec:5.3}
The fine-tuning process for proposed framework demonstrated significant improvement in the identification of OE for Nastaliq Urdu. However, its classification process is still a black-box and difficult to comprehend. As described earlier, model interpretability is a core objective of this research and equally important. Therefore, to make the decision-making inference understandable, we utilize the SHAP post-hoc model (\C{liu2024}{J. Liu et al., 2024}). It is a state-of-the-art method to estimate the contribution of each feature to the model outcome (\C{ali2023b}{S. Ali et al., 2023}). The explainability of the proposed model helps us to trust on the outcomes and make the system accountable.

\begin{figure}[!htbp]
\centering
\includegraphics[width=0.68\linewidth]{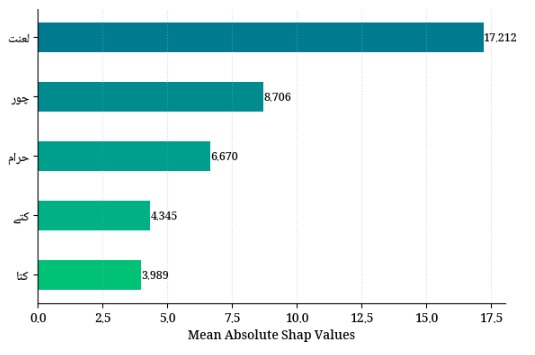}
\caption{Global SHAP bar plot of the top-5 contributing words across offensive test instances}
\label{fig:7}
\end{figure}

At first, we have estimated the top-5 features/words from the offensive instances (test part of the dataset), and SHAP bar plot is shown in Fig.~\ref{fig:7}. This plot presents the global feature importance based on the mean absolute SHAP values, that indicate each word/feature's overall contributions to the model's predictions. These top-5 features/words with corresponding SHAP value and translation in English is also presented in Table~\ref{tab:15}. It is evident that ``\ur{lanat}'' with a mean SHAP value of 17.212, has the strongest influence on the model's output, followed by the word ``\ur{chor}'' (8.706), which also plays a substantial role in driving predictions. The next two words ``\ur{haram}'' (6.670) and ``\ur{kutte}'' (4.345) demonstrate nearly equivalent contributions, reflecting moderate but consistent impact in driving predictions. Lastly, ``\ur{kutta}'' shows the lowest importance among the top-5 features. Overall, the first two features demonstrate noticeably stronger explanatory power, indicating their critical role in shaping the model's behavior.

\begin{table}[htbp]
\centering
\caption{Top-5 contributing features/words from offensive samples}
\label{tab:15}
\small
\renewcommand{\arraystretch}{1.5}
\begin{tabular}{@{}ccc>{\raggedright\arraybackslash}p{7cm}@{}}
\toprule
\textbf{Rank} & \textbf{Feature (Urdu)} & \textbf{\begin{tabular}[b]{@{}c@{}}Aggregate Absolute\\SHAP Contribution\end{tabular}} & \textbf{Translation / Context} \\
\midrule
1 & \ur{lanat} & 17.212 & Curse / Damnation (Strong Disapproval) \\
2 & \ur{chor} & 8.706 & Thief / Corrupt (Often used politically) \\
3 & \ur{haram} & 6.670 & Forbidden / Illegitimate (Serious religious/moral slur) \\
4 & \ur{kutte} & 4.345 & Dogs (Common derogatory plural slur) \\
5 & \ur{kutta} & 3.989 & Dog (Singular slur/Insult) \\
\bottomrule
\end{tabular}
\end{table}

\begin{figure}[!htbp]
\centering
\includegraphics[width=1.0\linewidth]{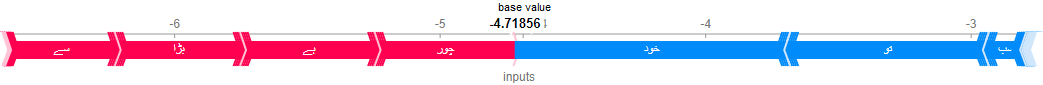}

\medskip
\includegraphics[width=\linewidth,height=1.2cm,keepaspectratio]{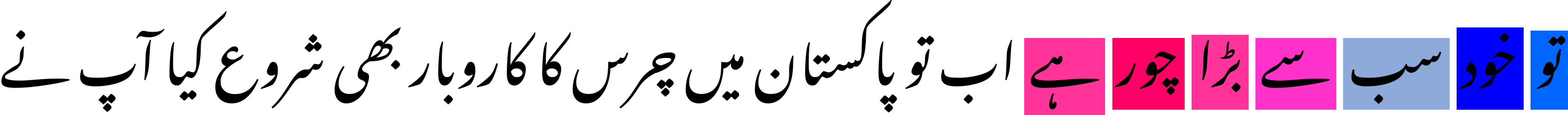}

\smallskip
{\small English Translation: You are the biggest thief yourself now you have also started a hashish business in Pakistan}
\caption{Offensive sample 1, SHAP force plot with base value highlighting important words in shades}
\label{fig:8}
\end{figure}

Next, we have selected four random samples in which three are offensive and one is non-offensive from the test dataset. First three samples are predicted as offensive by the proposed framework (MuRIL with fine-tuning). The SHAP force plot for sample 1 with shaded Nastaliq Urdu text is presented in Fig.~\ref{fig:8}. Each word receives a corresponding Shapley value, indicating its contribution to the prediction of OE. The shades of red colors for Urdu words indicate the intensity of positive contribution for predicting offensive class. Dark red represents higher whereas light red represents lower positive contribution. Likewise, shades of blue color for highlighted Urdu words indicate the negative contribution towards prediction.

\begin{figure}[!htbp]
\centering
\includegraphics[page=3,width=0.75\linewidth]{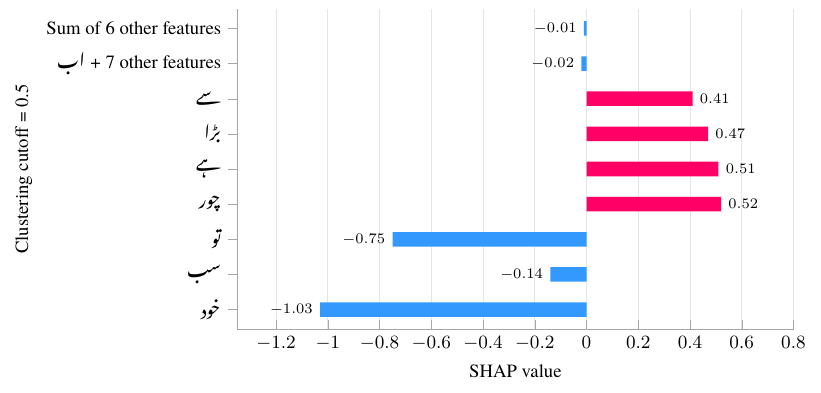}
\caption{SHAP bar plot containing Shapley values for words in Nastaliq Urdu OE [sample 1]}
\label{fig:9}
\end{figure}

The bar plot proposed by the SHAP method indicating the SHAP value for each word (sample 1) is presented in Fig.~\ref{fig:9}. It is evident that the word [\ur{chor}, thief] receives 0.52 SHAP value (receives dark red shade), which is highest positive contribution for prediction. This word [\ur{chor}, thief] is also selected among the top-5 (Fig.~\ref{fig:7}) features/words from offensive samples. Likewise, the word [\ur{bara}, big] receives 0.47, and the word [\ur{hai} , is] receives 0.51 SHAP values with corresponding red shades, indicating a strong positive contribution towards prediction. However, neither \ur{hai}, a copula occurring in most declarative sentences, nor \ur{bara} (big) carries offensive meaning. That a function word attracts attribution nearly equal to \ur{chor} suggests the model responds in part to distributional regularities of offensive comments rather than to offensive content alone. In contrast, the other words [\ur{khud}, himself], [\ur{tu}, you] receive blue shades, show negative contribution in prediction.

\begin{figure}[!htbp]
\centering
\includegraphics[width=1.0\linewidth]{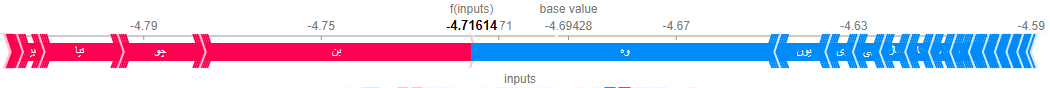}

\medskip
\includegraphics[width=\linewidth,height=1.2cm,keepaspectratio]{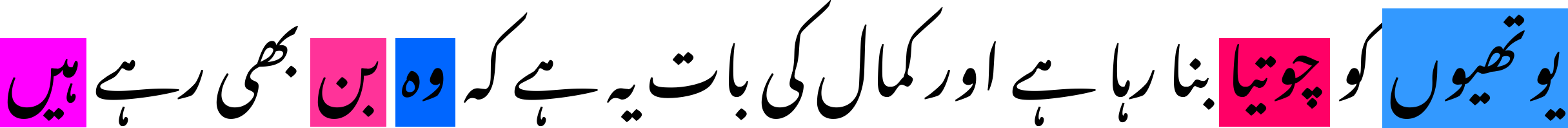}

\smallskip
{\small English Translation: He is making fools of the Youthias, and the remarkable thing is that they are becoming fools too}
\caption{Offensive sample 2, SHAP force plot with base value highlighting important words in shades}
\label{fig:10}
\end{figure}

Similarly, for sample 2, the SHAP force plot with highlighted Urdu words in red and blue shades is presented in Fig.~\ref{fig:10}. This sample is predicted as offensive by the proposed model. The word [\ur{chutiya}, moron] receives 0.04 SHAP value and the word [\ur{ban}, become] receives 0.06 SHAP value. These are typical words being used in offensive communciation. The words in blues shades [\ur{youthiyon}, Uthians] are contributing negatively in the prediction of offensive class.

\begin{figure}[!htbp]
\centering
\includegraphics[page=4,width=0.75\linewidth]{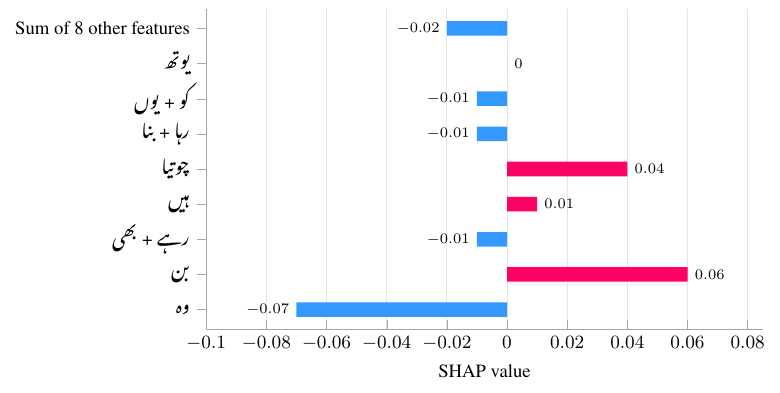}
\caption{SHAP bar plot containing Shapley values for words in Nastaliq Urdu OE [sample 2]}
\label{fig:11}
\end{figure}

The SHAP bar plot for sample 2 is presented in Fig.~\ref{fig:11}, indicating the SHAP value received by each significant Urdu word in sample 2. We can analyze the base value of -4.69 and the SHAP values of all words are summed into the base value. Thus, the SHAP explainable model has precisely interpreted the prediction of sample 2.

\begin{figure}[!htbp]
\centering
\includegraphics[width=1.0\linewidth]{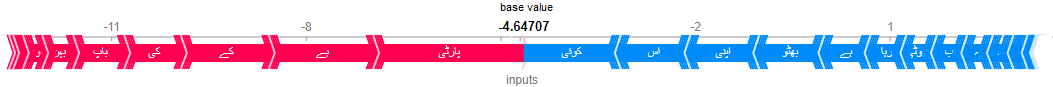}

\medskip
\includegraphics[width=\linewidth,height=1.2cm,keepaspectratio]{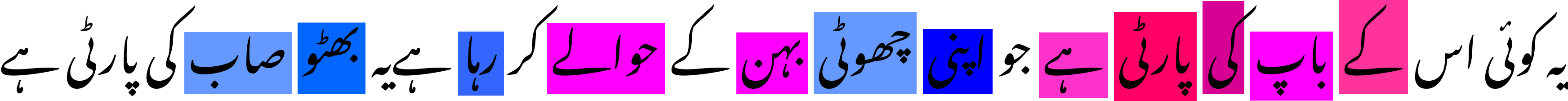}

\smallskip
{\small English Translation: This is his father's party who is handing over to his younger sister. This is Bhutto Saab's party}
\caption{Offensive sample 3, SHAP force plot with base value highlighting important words in shades}
\label{fig:12}
\end{figure}

\begin{figure}[!htbp]
\centering
\includegraphics[page=5,width=0.8\linewidth]{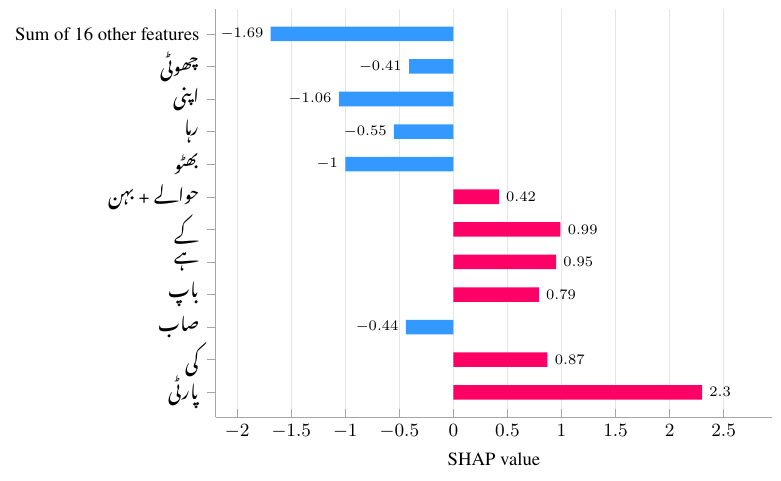}
\caption{SHAP bar plot containing Shapley values for words in Nastaliq Urdu OE [sample 3]}
\label{fig:13}
\end{figure}

Next, we took 3\textsuperscript{rd} sample to evaluate the effectiveness of the SHAP XAI model for the explainability of prediction. The SHAP force plot for sample 3 with base value is presented in Fig.~\ref{fig:12}. The positive words are highlighted with red shades, and strong words have dark shades, e.g. [\ur{party}, party], [\ur{baap}, father]. The words [\ur{apnichoti}] in blue color shades have a negative impact on the prediction. The SHAP bar plot for sample 3 is presented in Fig.~\ref{fig:13}, showing the contribution of each Urdu word toward prediction in the form of SHAP value. The words with red shades are often used in OEs to offend someone. In addition, it also highlights how an individual word's SHAP value affects the base value and corresponding prediction probability.

\begin{figure}[!htbp]
\centering
\includegraphics[width=1.0\linewidth]{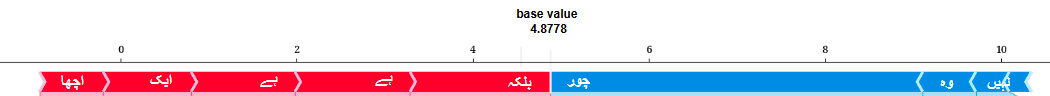}

\medskip
\includegraphics[width=\linewidth,height=1.2cm,keepaspectratio]{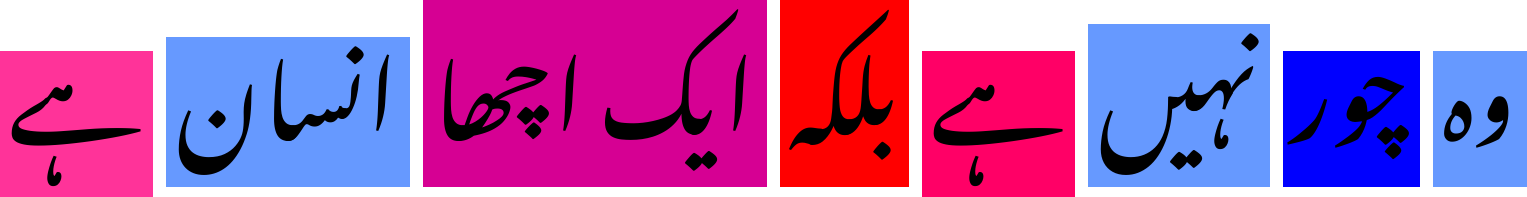}

\smallskip
{\small English Translation: He is not a thief, but a good person.}
\caption{Non-offensive sample 4, SHAP force plot with base value highlighting important words in shades}
\label{fig:14}
\end{figure}

Next, we considered 4th sample (non-offensive class) to evaluate the explanations provided by the SHAP model. The SHAP model predicted this sample as non-offensive and SHAP force plot with base value is presented in Fig.~\ref{fig:14}. Here, positive and negative words are highlighted by the corresponding colors. The word [\ur{chor}, thief] is most influential with a SHAP value of -4.3, but it affects negatively to the offensive class and positively to the non-offensive class. Although this word [\ur{chor}, thief] is the top-2 for offensive class but here the context is changed and it is being used for the negation. This is genuine contextual processing: the model has registered \ur{nahin} and inverted the contribution of the term it governs, which lexical matching alone cannot produce.

\begin{figure}[!htbp]
\centering
\includegraphics[width=0.9\linewidth]{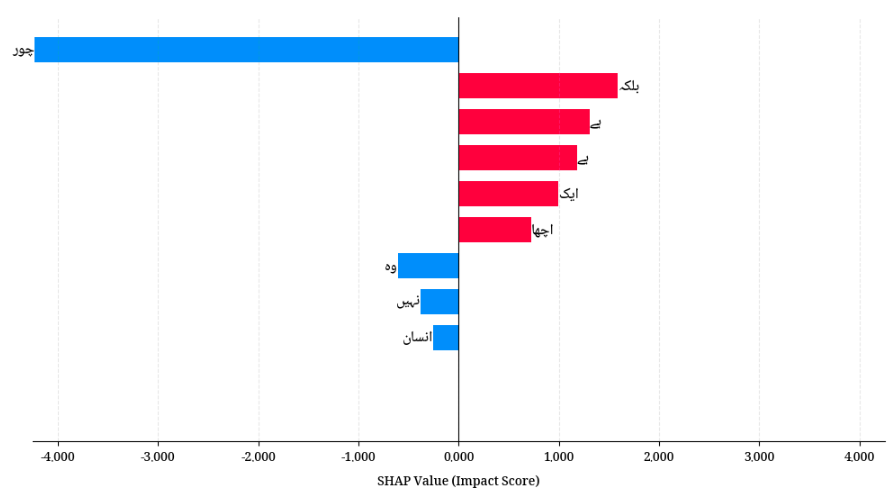}
\caption{SHAP bar plot containing SHAP values for words in Nastaliq Urdu OE [sample 4]}
\label{fig:15}
\end{figure}

Likewise the other words [\ur{woh}, that], [\ur{nahin} , Not] have relatively smaller positive impact on the prediction for non-offensive class. The SHAP bar plot for sample 4 is presented in Fig.~\ref{fig:15}. The word [\ur{balke}, Rather] carries the largest positive attribution toward the offensive class (1.6), but this is outweighed by the negative contribution of \ur{chor} (-4.3), yielding the non-offensive prediction overall.

The four instances indicate uneven competence. The model handles negation correctly (sample 4), which requires syntactic sensitivity. In three of four cases, however, substantial attribution falls on function words and topical vocabulary rather than on what makes each comment offensive. The global lexicon in Table~\ref{tab:15} is genuinely offensive; the instance-level evidence is not consistently so. Aggregate performance of 87.05\% may therefore rest partly on lexical and topical regularities specific to this corpus rather than on a transferable representation of offensiveness -- a distinction that accuracy and macro F1 cannot make, and one that bears directly on deployment across different pages, platforms, or political periods. Three caveats apply. Attribution to high-frequency function words may partly reflect the explanation method rather than the model, since SHAP depends on the background distribution and subword tokenization of Urdu requires aggregation to word level; separating the two would require perturbation-based validation, which we leave to future work. Four instances cannot establish prevalence, and a systematic study across the full test set would be needed to quantify how often attribution falls outside offensive vocabulary. Finally, the study does not evaluate whether the generated explanations are faithful to the model's decision process, which remains an open question for Nastaliq Urdu.

\section{Discussion}\label{sec:6}
The curation reported in Section~\ref{sec:3.3} has implications beyond this study. Independent re-annotation by three annotators who had not seen the original labels resulted in a class change for 500 comments, approximately 6.6\% of the corpus, with inter-annotator agreement rising from 0.88 to 0.90. A revision rate of this magnitude in a corpus that had already been annotated by three experienced native speakers and published suggests that label noise in low-resource offensive-language resources may be more substantial than reported agreement scores alone convey. Offensiveness judgements depend on context, register, and the annotator's own social position, and a kappa of 0.88 is consistent with a non-trivial number of borderline cases being resolved differently by different annotators. For downstream users, this means that reported performance on such corpora carries an uncertainty attributable to the labels themselves, independent of model quality. We therefore release the curated corpus with its annotation guidelines and datasheet, so that subsequent work can assess these judgements directly rather than inheriting them.

While class balancing is a widely adopted technique to prevent model bias toward the majority class during training, it may introduce limitations when applied to tasks such as OED, where real-world data is inherently imbalanced. On social media platforms like Facebook, offensive content represents only a small fraction of all user-generated posts in the natural settings. As a result, the real-world class distribution is heavily skewed, often with far more non-offensive samples than offensive ones. Training a model on a balanced 1:1 dataset may artificially amplify the prevalence of offensive content, which can lead to several generalization issues:

\textbf{Misleading Performance Metrics:} Metrics such as accuracy computed on a balanced test set may not reflect the model's true performance in deployment, where the class imbalance can result in higher false positive rates. A model trained on balanced data may learn to over-predict the minority class (offensive), which can degrade user experience by flagging non-offensive content unnecessarily.

\textbf{Sensitivity to Class Priors:} By training with equal class proportions, the model becomes less sensitive to the true prior probability of each class. As a result, the model's decision boundary may not be optimal when applied to naturally skewed data, this will lead to reduced effectiveness in real-world applications.

\textbf{Overfitting to Minority Class Patterns:} In 1:1 sampling, minority class samples are used repeatedly to match the majority class, which may lead to overfitting. The model might learn patterns that are specific to the training set but not generalizable to unseen, naturally distributed data.

\textbf{Impact on Deployment-Phase Trade-offs:} Models trained under balanced conditions may require additional calibration when deployed, such as threshold tuning or the use of probability scaling techniques. Without such adjustments, the system may misclassify a large number of negative instances as positive.

On the other hand, the current study focuses on the detection of offensive expressions in Facebook comments written in Nastaliq Urdu. Facebook was selected as the source platform due to the availability of public comments in the target script and its widespread use among Urdu-speaking communities. However, the underlying architecture of the proposed framework is designed to be platform-agnostic and adaptable to a wide range of textual data sources.

Technically, the adaptability of this framework relies on its modular preprocessing pipeline, which includes steps such as language normalization, Nastaliq script handling, tokenization, and domain-specific vocabulary augmentation. These components can be easily reconfigured to process user-generated content from other platforms like Twitter, YouTube comments, or Reddit, where linguistic style, post length, and noise levels may differ. With appropriate data curation and platform-specific noise reduction techniques (e.g., handling hashtags and emojis on Twitter or comment threading on YouTube), the model can be retrained or fine-tuned to accommodate the nuances of each domain.

Additionally, the transformer models used (e.g., MuRIL, Twhin-BERT) are pre-trained on multilingual or social media-specific corpora, making them well-suited for domain adaptation tasks. Fine-tuning these models on new platform-specific data, while reapplying SHAP for interpretability, would allow us to maintain the explainability of predictions even in shifted domains. Our evaluation framework, including MARS Shine-through and Occlusion metrics, is also generalizable and can be reused to assess model robustness and uniqueness in new contexts. As part of our future work, we aim to extend the dataset by collecting and annotating comments from other social media platforms and evaluate the cross-domain transferability of the proposed framework.

Although our model performs well on Nastaliq Urdu from Facebook, we acknowledge its current focus as a potential limitation regarding the broader Urdu digital forms. Social media often features Naskh script and Roman Urdu, alongside code-mixing (Urdu-English), which introduces structural and lexical complexities, not covered by our dataset. The model was trained exclusively on Nastaliq-script text, so performance on Naskh or Roman Urdu input cannot be inferred from the results reported here, and degradation on multi-script or code-mixed content is likely. Future research will prioritize expanding the dataset to include these variants of script and try to integrate cross-lingual embeddings to improve the framework's robustness across the diverse styles of Urdu language.

\section{Conclusion and future work}\label{sec:7}
In this study, we addressed offensive expression detection in monolingual Nastaliq Urdu Facebook comments through three contributions. First, we curated and publicly released a corpus of 7,604 comments, extending an existing Facebook corpus through independent re-annotation that changed the class assigned to 6.6\% of instances and raised inter-annotator agreement from 0.88 to 0.90. Second, we fine-tuned four Urdu-capable transformers under a common grid-search protocol and evaluated them against twelve feature--classifier baselines on identical partitions. MuRIL achieved the highest scores at 87.05\% accuracy and 87.06\% macro F1, with performance stable across five optimizers and the leading two transformers separated by less than the test set can resolve, with 95\% confidence intervals on accuracy spanning approximately $\pm$1.7 percentage points. MARS Shine-Through and Occlusion indicated that the fine-tuned models recover complementary subsets of offensive instances. Third, word-level SHAP attributions adapted to right-to-left rendering showed that while the model has acquired a lexicon of offensive terms and handles negation correctly, substantial attribution also falls on syntactically neutral function words and on political vocabulary characteristic of the source pages. Aggregate performance on this task may therefore depend in part on lexical and topical regularities rather than on a transferable representation of offensiveness --- a distinction that accuracy and F1 alone cannot make, and one that should inform any consideration of deployment.

This work is limited to monolingual, single-script (Nastaliq), single-platform (Facebook) data, and the attribution analysis covers four instances rather than the full test set. Future work will quantify the prevalence of lexical shortcut reliance across the corpus, evaluate robustness under script-level perturbation, extend the framework to multi-script and code-mixed Urdu, and assess cross-platform transfer. We also intend to evaluate the faithfulness of the generated explanations, which the present study does not establish.

\section*{Data Availability}\sloppy

The curated Nastaliq Urdu offensive expression corpus, together with its annotation guidelines and datasheet, is publicly available at: \url{https://github.com/MSIMALIK/Offensive-Expression-detection-In-Urdu}

The code of the proposed framework is available at the same repository.

The code for Adam optimizer is available at: \url{https://github.com/theroyakash/Adam}

The code for AdaBelief optimizer is available at: \url{https://github.com/juntang-zhuang/Adabelief-Optimizer}

The code for AdaBound optimizer is available at: \url{https://github.com/Luolc/AdaBound}

The code for AdaBoB optimizer is available at: \url{https://gitee.com/frontxiang/torch_adabob}

\section*{References}
\begingroup\small\setlength{\parindent}{0pt}\setlength{\parskip}{3pt}
\hangindent=1.5em\hangafter=1\hypertarget{ref.abbas2023}{}Abbas, Y., \& Malik, M. S. I. (2023). Defective products identification framework using online reviews. \emph{Electronic Commerce Research, 23}(2), 899--920.\par
\hangindent=1.5em\hangafter=1\hypertarget{ref.akhter2020}{}Akhter, M. P., Jiangbin, Z., Naqvi, I. R., Abdelmajeed, M., \& Sadiq, M. T. (2020). Automatic detection of offensive language for urdu and roman urdu. \emph{IEEE access, 8}, 91213--91226.\par
\hangindent=1.5em\hangafter=1\hypertarget{ref.akhter2021}{}Akhter, M. P., Jiangbin, Z., Naqvi, I. R., AbdelMajeed, M., \& Zia, T. (2021). Abusive language detection from social media comments using conventional machine learning and deep learning approaches. \emph{Multimedia Systems}, 1--16.\par
\hangindent=1.5em\hangafter=1\hypertarget{ref.ali2023a}{}Ali, G., \& Malik, M. S. I. (2023). Rumour identification on Twitter as a function of novel textual and language-context features. \emph{Multimedia Tools and Applications, 82}(5), 7017--7038.\par
\hangindent=1.5em\hangafter=1\hypertarget{ref.ali2023b}{}Ali, S., Abuhmed, T., El-Sappagh, S., Muhammad, K., Alonso-Moral, J. M., Confalonieri, R., . . . Herrera, F. (2023). Explainable Artificial Intelligence (XAI): What we know and what is left to attain Trustworthy Artificial Intelligence. \emph{Information fusion, 99}, 101805.\par
\hangindent=1.5em\hangafter=1\hypertarget{ref.amjad2021}{}Amjad, M., Ashraf, N., Sidorov, G., Zhila, A., Chanona-Hernandez, L., \& Gelbukh, A. (2021). Automatic abusive language detection in urdu tweets. \emph{Acta Polytechnica Hungarica}, 1785--8860.\par
\hangindent=1.5em\hangafter=1\hypertarget{ref.ashraf2024}{}Ashraf, K., Hosen, M. H., Asgar, S., Islam, M. T., \& Nawar, S. (2024). \emph{Analyzing abusive bangla comments on social media: Nlp \& explainable ai.} Paper presented at the 2024 International Conference on Advances in Computing, Communication, Electrical, and Smart Systems (iCACCESS).\par
\hangindent=1.5em\hangafter=1\hypertarget{ref.ayetiran2024}{}Ayetiran, E. F., \& Özgöbek, Ö. (2024). An inter-modal attention-based deep learning framework using unified modality for multimodal fake news, hate speech and offensive language detection. \emph{Information Systems, 123}, 102378.\par
\hangindent=1.5em\hangafter=1\hypertarget{ref.bennetot2022}{}Bennetot, A., Franchi, G., Del Ser, J., Chatila, R., \& Diaz-Rodriguez, N. (2022). Greybox XAI: A Neural-Symbolic learning framework to produce interpretable predictions for image classification. \emph{Knowledge-Based Systems, 258}, 109947.\par
\hangindent=1.5em\hangafter=1\hypertarget{ref.chen2019}{}Chen, C., Li, O., Tao, D., Barnett, A., Rudin, C., \& Su, J. K. (2019). This looks like that: deep learning for interpretable image recognition. \emph{Advances in neural information processing systems, 32}.\par
\hangindent=1.5em\hangafter=1\hypertarget{ref.chen2017}{}Chen, H., Mckeever, S., \& Delany, S. J. (2017). \emph{Harnessing the power of text mining for the detection of abusive content in social media.} Paper presented at the Advances in Computational Intelligence Systems: Contributions Presented at the 16th UK Workshop on Computational Intelligence, September 7--9, 2016, Lancaster, UK.\par
\hangindent=1.5em\hangafter=1\hypertarget{ref.coltekin2020}{}Çöltekin, Ç. (2020). \emph{A corpus of Turkish offensive language on social media.} Paper presented at the Proceedings of the Twelfth Language Resources and Evaluation Conference.\par
\hangindent=1.5em\hangafter=1\hypertarget{ref.dandolo2023}{}Dandolo, D., Masiero, C., Carletti, M., Dalle Pezze, D., \& Susto, G. A. (2023). AcME---Accelerated model-agnostic explanations: Fast whitening of the machine-learning black box. \emph{Expert Systems with Applications, 214}, 119115.\par
\hangindent=1.5em\hangafter=1\hypertarget{ref.daud2015}{}Daud, A., Ahmad, M., Malik, M., \& Che, D. (2015). Using machine learning techniques for rising star prediction in co-author network. \emph{Scientometrics, 102}, 1687--1711.\par
\hangindent=1.5em\hangafter=1\hypertarget{ref.davidson2017}{}Davidson, T., Warmsley, D., Macy, M., \& Weber, I. (2017). \emph{Automated hate speech detection and the problem of offensive language.} Paper presented at the Proceedings of the international AAAI conference on web and social media.\par
\hangindent=1.5em\hangafter=1\hypertarget{ref.devlin2018}{}Devlin, J., Chang, M.-W., Lee, K., \& Toutanova, K. (2018). Bert: Pre-training of deep bidirectional transformers for language understanding. \emph{arXiv preprint \href{https://arxiv.org/abs/1810.04805}{arXiv:1810.04805}}.\par
\hangindent=1.5em\hangafter=1\hypertarget{ref.fetahi2025}{}Fetahi, E., Susuri, A., Hamiti, M., Kastrati, Z., Canhasi, E., \& Misini, A. (2025). Enhancing social media hate speech detection in low-resource languages using transformers and explainable AI. \emph{Social Network Analysis and Mining, 15}(1), 82.\par
\hangindent=1.5em\hangafter=1\hypertarget{ref.hashir2025}{}Hashir, M. H., \& Kim, S. W. (2025). TARGE: large language model-powered explainable hate speech detection. \emph{PeerJ Computer Science, 11}, e2911.\par
\hangindent=1.5em\hangafter=1\hypertarget{ref.hashmi2025}{}Hashmi, E., Yildirim Yayilgan, S., Abomhara, M., \& Akerkar, R. (2025). \emph{Transparent Hate Speech Detection in Norwegian Using Explainable AI.} Paper presented at the International Conference on Intelligent and Fuzzy Systems.\par
\hangindent=1.5em\hangafter=1\hypertarget{ref.hinduja2010}{}Hinduja, S., \& Patchin, J. W. (2010). Bullying, cyberbullying, and suicide. \emph{Archives of suicide research, 14}(3), 206--221.\par
\hangindent=1.5em\hangafter=1\hypertarget{ref.hu2020}{}Hu, J., Ruder, S., Siddhant, A., Neubig, G., Firat, O., \& Johnson, M. (2020). \emph{Xtreme: A massively multilingual multi-task benchmark for evaluating cross-lingual generalisation.} Paper presented at the International Conference on Machine Learning.\par
\hangindent=1.5em\hangafter=1\hypertarget{ref.humayoun2022}{}Humayoun, M. (2022). Abusive and threatening language detection in urdu using supervised machine learning and feature combinations. \emph{arXiv preprint \href{https://arxiv.org/abs/2204.03062}{arXiv:2204.03062}}.\par
\hangindent=1.5em\hangafter=1\hypertarget{ref.hussain2022}{}Hussain, S., Malik, M. S. I., \& Masood, N. (2022). Identification of offensive language in Urdu using semantic and embedding models. \emph{PeerJ Computer Science, 8}, e1169.\par
\hangindent=1.5em\hangafter=1\hypertarget{ref.jahan2021}{}Jahan, M. S., \& Oussalah, M. (2021). A systematic review of Hate Speech automatic detection using natural language processing. \emph{arXiv preprint \href{https://arxiv.org/abs/2106.00742}{arXiv:2106.00742}}.\par
\hangindent=1.5em\hangafter=1\hypertarget{ref.jung2017}{}Jung, J., Concannon, C., Shroff, R., Goel, S., \& Goldstein, D. G. (2017). Simple rules for complex decisions. \emph{arXiv preprint \href{https://arxiv.org/abs/1702.04690}{arXiv:1702.04690}}.\par
\hangindent=1.5em\hangafter=1\hypertarget{ref.khan2024}{}Khan, M. S., Malik, M. S. I., \& Nadeem, A. (2024). Detection of violence incitation expressions in Urdu tweets using convolutional neural network. \emph{Expert Systems with Applications}, 123174.\par
\hangindent=1.5em\hangafter=1\hypertarget{ref.kumar2021}{}Kumar, A., Saumya, S., \& Roy, P. K. (2021). \emph{Abusive and threatening language detection from Urdu social media posts: A machine learning approach.} Paper presented at the Working Notes of FIRE 2021-Forum for Information Retrieval Evaluation (Online). CEUR.\par
\hangindent=1.5em\hangafter=1\hypertarget{ref.liu2024}{}Liu, J., Wu, X., Xie, Y., Tang, Z., Xie, Y., \& Gong, S. (2024). Small samples-oriented intrinsically explainable machine learning using Variational Bayesian Logistic Regression: An intensive care unit readmission prediction case for liver transplantation patients. \emph{Expert Systems with Applications, 235}, 121138.\par
\hangindent=1.5em\hangafter=1\hypertarget{ref.liu2019}{}Liu, Y., Ott, M., Goyal, N., Du, J., Joshi, M., Chen, D., . . . Stoyanov, V. (2019). Roberta: A robustly optimized bert pretraining approach. \emph{arXiv preprint \href{https://arxiv.org/abs/1907.11692}{arXiv:1907.11692}}.\par
\hangindent=1.5em\hangafter=1\hypertarget{ref.lundberg2017}{}Lundberg, S. M., \& Lee, S.-I. (2017). A unified approach to interpreting model predictions. \emph{Advances in neural information processing systems, 30}.\par
\hangindent=1.5em\hangafter=1\hypertarget{ref.malik2020b}{}Malik, M. S. I. (2020). Predicting users' review helpfulness: the role of significant review and reviewer characteristics. \emph{Soft Computing, 24}(18), 13913--13928.\par
\hangindent=1.5em\hangafter=1\hypertarget{ref.malik2023a}{}Malik, M. S. I. (2023). \emph{Threatening Expression and Target Identification in Under-Resource Languages Using NLP Techniques.} Paper presented at the International Conference on Analysis of Images, Social Networks and Texts.\par
\hangindent=1.5em\hangafter=1\hypertarget{ref.malik2024a}{}Malik, M. S. I., \& Nawaz, A. (2024). SEHP: stacking-based ensemble learning on novel features for review helpfulness prediction. \emph{Knowledge and Information Systems, 66}(1), 653--679.\par
\hangindent=1.5em\hangafter=1\hypertarget{ref.malik2023b}{}Malik, M. S. I., Cheema, U., \& Ignatov, D. I. (2023). Contextual Embeddings based on Fine-tuned Urdu-BERT for Urdu Threatening Content and Target Identification. \emph{Journal of King Saud University-Computer and Information Sciences}, 101606.\par
\hangindent=1.5em\hangafter=1\hypertarget{ref.malik2023c}{}Malik, M. S. I., Imran, T., \& Mamdouh, J. M. (2023). How to detect propaganda from social media? Exploitation of semantic and fine-tuned language models. \emph{PeerJ Computer Science, 9}, e1248.\par
\hangindent=1.5em\hangafter=1\hypertarget{ref.malik2023d}{}Malik, M. S. I., Nawaz, A., Jamjoom, M. M., \& Ignatov, D. I. (2023). Effectiveness of ELMo embeddings, and semantic models in predicting review helpfulness. \emph{Intelligent Data Analysis}(Preprint), 1--21.\par
\hangindent=1.5em\hangafter=1\hypertarget{ref.malik2025}{}Malik, M. S. I., Nazarova, A., \& Jamjoom, M. M. (2025). Explainable threatening tweets identification using abstract meaning representation graph and pattern structure approaches. \emph{Intelligent Data Analysis}, 1088467X251330977.\par
\hangindent=1.5em\hangafter=1\hypertarget{ref.malik2023e}{}Malik, M. S. I., Nazarova, A., Jamjoom, M. M., \& Ignatov, D. I. (2023). Multilingual hope speech detection: A Robust framework using transfer learning of fine-tuning RoBERTa model. \emph{Journal of King Saud University-Computer and Information Sciences, 35}(8), 101736.\par
\hangindent=1.5em\hangafter=1\hypertarget{ref.malik2024b}{}Malik, M. S. I., Rehman, F., \& Ignatov, D. I. (2024). Ensemble learning with linguistic, summary language and psychological features for location prediction. \emph{International Journal of Information Technology, 16}(1), 193--205.\par
\hangindent=1.5em\hangafter=1\hypertarget{ref.malik2024c}{}Malik, M. S. I., Younas, M. Z., Jamjoom, M. M., \& Ignatov, D. I. (2024). Categorization of tweets for damages: infrastructure and human damage assessment using fine-tuned BERT model. \emph{PeerJ Computer Science, 10}, e1859.\par
\hangindent=1.5em\hangafter=1\hypertarget{ref.malik2017}{}Malik, M., \& Hussain, A. (2017). Helpfulness of product reviews as a function of discrete positive and negative emotions. \emph{Computers in Human Behavior, 73}, 290--302.\par
\hangindent=1.5em\hangafter=1\hypertarget{ref.malik2018}{}Malik, M., \& Hussain, A. (2018). An analysis of review content and reviewer variables that contribute to review helpfulness. \emph{Information Processing \& Management, 54}(1), 88--104.\par
\hangindent=1.5em\hangafter=1\hypertarget{ref.malik2020a}{}Malik, M., \& Hussain, A. (2020). Exploring the influential reviewer, review and product determinants for review helpfulness. \emph{Artificial Intelligence Review, 53}, 407--427.\par
\hangindent=1.5em\hangafter=1\hypertarget{ref.mazari2024}{}Mazari, A. C., Boudoukhani, N., \& Djeffal, A. (2024). BERT-based ensemble learning for multi-aspect hate speech detection. \emph{Cluster Computing, 27}(1), 325--339.\par
\hangindent=1.5em\hangafter=1\hypertarget{ref.mehboob2021}{}Mehboob, A., \& Malik, M. (2021). Smart fraud detection framework for job recruitments. \emph{Arabian Journal for Science and Engineering, 46}(4), 3067--3078.\par
\hangindent=1.5em\hangafter=1\hypertarget{ref.mubarak2020}{}Mubarak, H., Rashed, A., Darwish, K., Samih, Y., \& Abdelali, A. (2020). Arabic offensive language on twitter: Analysis and experiments. \emph{arXiv preprint \href{https://arxiv.org/abs/2004.02192}{arXiv:2004.02192}}.\par
\hangindent=1.5em\hangafter=1\hypertarget{ref.nawaz2022}{}Nawaz, A., \& Malik, M. S. I. (2022). Rising stars prediction in reviewer network. \emph{Electronic Commerce Research, 22}(1), 53--75.\par
\hangindent=1.5em\hangafter=1\hypertarget{ref.nazarova2024}{}Nazarova, A., Malik, M. S. I., Ignatov, D. I., \& Hussain, I. (2024). Deepthreatexplainer: a united explainable predictor for threat comments identification on Twitter. \emph{Social Network Analysis and Mining, 14}(1), 228.\par
\hangindent=1.5em\hangafter=1\hypertarget{ref.nghiem2024}{}Nghiem, H., \& Daumé III, H. (2024). Hatecot: An explanation-enhanced dataset for generalizable offensive speech detection via large language models. \emph{arXiv preprint \href{https://arxiv.org/abs/2403.11456}{arXiv:2403.11456}}.\par
\hangindent=1.5em\hangafter=1\hypertarget{ref.paaschcolberg2021}{}Paasch-Colberg, S., Strippel, C., Trebbe, J., \& Emmer, M. (2021). From insult to hate speech: Mapping offensive language in German user comments on immigration. \emph{Media and Communication, 9}(1), 171--180.\par
\hangindent=1.5em\hangafter=1\hypertarget{ref.papernot2018}{}Papernot, N., \& McDaniel, P. (2018). Deep k-nearest neighbors: Towards confident, interpretable and robust deep learning. \emph{arXiv preprint \href{https://arxiv.org/abs/1803.04765}{arXiv:1803.04765}}.\par
\hangindent=1.5em\hangafter=1\hypertarget{ref.pitenis2020}{}Pitenis, Z., Zampieri, M., \& Ranasinghe, T. (2020). Offensive language identification in Greek. \emph{arXiv preprint \href{https://arxiv.org/abs/2003.07459}{arXiv:2003.07459}}.\par
\hangindent=1.5em\hangafter=1\hypertarget{ref.raza2023}{}Raza, M. O., Mahoto, N. A., Hamdi, M., Al Reshan, M. S., Rajab, A., \& Shaikh, A. (2023). Detection of offensive terms in resource-poor language using machine learning algorithms. \emph{PeerJ Computer Science, 9}, e1524.\par
\hangindent=1.5em\hangafter=1\hypertarget{ref.rehan2023}{}Rehan, M., Malik, M. S. I., \& Jamjoom, M. M. (2023). Fine-tuning transformer models using transfer learning for multilingual threatening text identification. \emph{IEEE Access}.\par
\hangindent=1.5em\hangafter=1\hypertarget{ref.restrepo2022}{}Restrepo, F., Mali, N., Abrahams, A., \& Ractham, P. (2022). Formal definition of the MARS method for quantifying the unique target class discoveries of selected machine classifiers. \emph{F1000Research, 11}.\par
\hangindent=1.5em\hangafter=1\hypertarget{ref.ribeiro2016}{}Ribeiro, M. T., Singh, S., \& Guestrin, C. (2016). \emph{``Why should I trust you?'' Explaining the predictions of any classifier.} Paper presented at the Proceedings of the 22nd ACM SIGKDD international conference on knowledge discovery and data mining.\par
\hangindent=1.5em\hangafter=1\hypertarget{ref.roy2022}{}Roy, P. K., Bhawal, S., \& Subalalitha, C. N. (2022). Hate speech and offensive language detection in Dravidian languages using deep ensemble framework. \emph{Computer Speech \& Language, 75}, 101386.\par
\hangindent=1.5em\hangafter=1\hypertarget{ref.rudin2022}{}Rudin, C., Chen, C., Chen, Z., Huang, H., Semenova, L., \& Zhong, C. (2022). Interpretable machine learning: Fundamental principles and 10 grand challenges. \emph{Statistic Surveys, 16}, 1--85.\par
\hangindent=1.5em\hangafter=1\hypertarget{ref.saeed2023}{}Saeed, R., Afzal, H., Rauf, S. A., \& Iltaf, N. (2023). Detection of Offensive Language and ITS Severity for Low Resource Language. \emph{ACM Transactions on Asian and Low-Resource Language Information Processing, 22}(6), 1--27.\par
\hangindent=1.5em\hangafter=1\hypertarget{ref.saisubramanian2020}{}Saisubramanian, S., Galhotra, S., \& Zilberstein, S. (2020). \emph{Balancing the tradeoff between clustering value and interpretability.} Paper presented at the Proceedings of the AAAI/ACM Conference on AI, Ethics, and Society.\par
\hangindent=1.5em\hangafter=1\hypertarget{ref.schmitz1999}{}Schmitz, G. P., Aldrich, C., \& Gouws, F. S. (1999). ANN-DT: an algorithm for extraction of decision trees from artificial neural networks. \emph{IEEE Transactions on Neural Networks, 10}(6), 1392--1401.\par
\hangindent=1.5em\hangafter=1\hypertarget{ref.sigurbergsson2019}{}Sigurbergsson, G. I., \& Derczynski, L. (2019). Offensive language and hate speech detection for Danish. \emph{arXiv preprint \href{https://arxiv.org/abs/1908.04531}{arXiv:1908.04531}}.\par
\hangindent=1.5em\hangafter=1\hypertarget{ref.xiang2025}{}Xiang, Q., Wang, X., Lei, L., \& Song, Y. (2025). Dynamic bound adaptive gradient methods with belief in observed gradients. \emph{Pattern Recognition, 168}, 111819.\par
\hangindent=1.5em\hangafter=1\hypertarget{ref.xiao2024}{}Xiao, Y., Bouamor, H., \& Zaghouani, W. (2024). Chinese Offensive Language Detection: Current Status and Future Directions. \emph{arXiv preprint \href{https://arxiv.org/abs/2403.18314}{arXiv:2403.18314}}.\par
\hangindent=1.5em\hangafter=1\hypertarget{ref.younas2023}{}Younas, M. Z., Malik, M. S. I., \& Ignatov, D. I. (2023). Automated defect identification for cell phones using language context, linguistic and smoke-word models. \emph{Expert Systems with Applications, 227}, 120236.\par
\hangindent=1.5em\hangafter=1\hypertarget{ref.zhang2018}{}Zhang, Q., Wu, Y. N., \& Zhu, S.-C. (2018). \emph{Interpretable convolutional neural networks.} Paper presented at the Proceedings of the IEEE conference on computer vision and pattern recognition.\par
\hangindent=1.5em\hangafter=1\hypertarget{ref.zhang2022}{}Zhang, X., Malkov, Y., Florez, O., Park, S., McWilliams, B., Han, J., \& El-Kishky, A. (2022). TwHIN-BERT: a socially-enriched pre-trained language model for multilingual Tweet representations. \emph{arXiv preprint \href{https://arxiv.org/abs/2209.07562}{arXiv:2209.07562}}.\par
\endgroup

\end{document}